\documentclass[10pt,twocolumn,letterpaper]{article}
\usepackage[pagenumbers]{cvpr}
\usepackage{seqsplit}
\usepackage{xurl}
\definecolor{carryqprimary}{HTML}{8080FF}
\usepackage[pagebackref,breaklinks,colorlinks,allcolors=carryqprimary]{hyperref}
\author{%
Shuaijun Liu\quad Feiyang You\quad Chenglong Zhang\\
Shuyang Hao\quad Chengyu Wu\quad Ningxin Su\textsuperscript{*}\\[3pt]
{\normalsize The Hong Kong University of Science and Technology (Guangzhou)}\\
{\normalsize \textsuperscript{*}Corresponding author: \href{mailto:ningxinsu@hkust-gz.edu.cn}{ningxinsu@hkust-gz.edu.cn}}\\[2pt]
{\normalsize Project page: \href{https://nebulis-lab.com/CarryQ/}{https://nebulis-lab.com/CarryQ/}}%
}

\hypersetup{pdftitle={CarryQ: Temporal Error Compensation for Robot Action Tokenization},pdfauthor={Shuaijun Liu, Feiyang You, Chenglong Zhang, Shuyang Hao, Chengyu Wu, Ningxin Su}}
\usepackage{tabularx}
\usepackage{makecell}
\usepackage{colortbl}
\colorlet{tableaccent}{carryqprimary}
\colorlet{tablelight}{carryqprimary!24!white}
\colorlet{tablefaint}{carryqprimary!9!white}
\definecolor{tablegroupgray}{gray}{0.94}
\newcolumntype{Y}{>{\raggedright\arraybackslash}X}
\newcolumntype{Q}{>{\columncolor{tablefaint}}r}
\newcommand{\tablestyle}{\fontsize{8.5}{10}\selectfont\setlength{\tabcolsep}{3pt}%
  \setlength{\aboverulesep}{0pt}\setlength{\belowrulesep}{0pt}%
  \setlength{\abovetopsep}{0pt}\setlength{\belowbottomsep}{0pt}%
  \renewcommand{\arraystretch}{1.25}}
\newcommand{\tablesubfont}{\fontsize{6.8}{8}\selectfont}
\newcommand{\tablehead}{\rowcolor{tablelight}}
\newcommand{\tablegroup}[2]{\rowcolor{tablegroupgray}\multicolumn{#1}{l}{\strut\textit{#2}}\\}
\newcommand{\tablepair}[2]{\makecell[r]{#1\\{\tablesubfont\color{tableaccent} #2}}}
\newcommand{\tablelabel}[2]{\makecell[l]{#1\\{\tablesubfont\color{tableaccent} #2}}}
\newcommand{\cqhead}{\textcolor{tableaccent}{CarryQ}}

\newcommand{\gainbar}[1]{\begingroup\dimen0=#1pt\dimen0=.45\dimen0%
  \makebox[18pt][l]{\rlap{\textcolor{tablefaint}{\rule{18pt}{2.4pt}}}%
  \textcolor{tableaccent}{\rule{\dimen0}{2.4pt}}}\endgroup}
\newlength{\gainnumberwidth}
\newcommand{\gainvalue}[1]{\begingroup
  \settowidth{\gainnumberwidth}{\textbf{+00.00}}%
  \gainbar{#1}\hspace{4pt}\makebox[\gainnumberwidth][r]{\textbf{+#1}}%
  \endgroup}
\newcommand{\pairedci}[1]{{\tablesubfont\color{tableaccent} CI [#1]}}
\newcommand{\countchange}[3]{\tablepair{#1 $\rightarrow$ #2}{successes / #3}}
\newcommand{\tablenoterow}[3]{\midrule
  \multicolumn{#1}{p{\dimexpr#2-2\tabcolsep\relax}}{%
  \fontsize{7.5}{9}\selectfont\raggedright\strut #3\strut}\tabularnewline
  \bottomrule}
\newcommand{\inlineci}[2]{#1\enspace{\tablesubfont\color{tableaccent}[#2]}}
\newcommand{\inlinecount}[2]{#1\enspace{\tablesubfont\color{tableaccent}(#2)}}

\title{CarryQ: Temporal Error Compensation for Robot Action Tokenization}

\usepackage{stfloats}

\begin{document}

\maketitle
\begin{figure*}[!b]
\centering
\includegraphics[width=\textwidth]{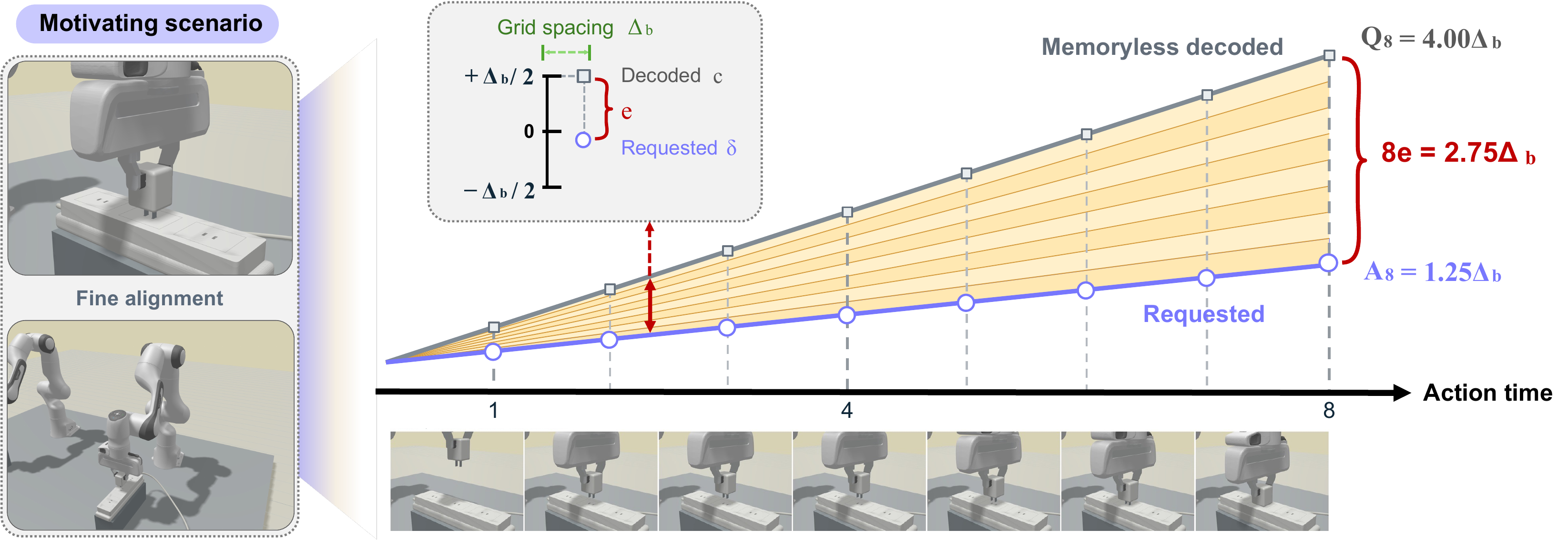}
\caption{Temporal quantization bias in fine alignment. Repeated rounding
from $\delta$ to $c$ accumulates error $e=c-\delta$.
$A_8$ and $Q_8$ are requested and decoded eight-step command sums.
Section~\ref{sec:observed-bias} examines this effect in recorded commands.}
\label{fig:motivation}
\end{figure*}

\begin{abstract}
Accurately reconstructing individual robot actions does not necessarily preserve the motion commanded by their sequence.
In fine manipulation, small quantization errors can reinforce across successive movements, producing substantial cumulative discrepancies that pointwise accuracy fails to capture.
We investigate this temporal error structure in robot action tokenization and introduce CarryQ, a residual-feedback encoder that enables later token choices to compensate for earlier rounding errors without changing the codebook, decoder, or token budget.
Our error decomposition reveals how stronger cross-time cancellation reduces cumulative command error even when pointwise distortion increases.
Across scalar, vector-codebook, and native Praxis tokenization, CarryQ improves matched-budget replay success by 8.84--30.67 percentage points, with further gains observed across closed-loop robot policies.
At 8 bits, mean simulated USB-C and charger insertion success improves from 15.42\% to 43.33\%.
On a physical robot, completed charger connections increase from 6\% to 36\% at 4 bits and from 16\% to 42\% at 6 bits.
These findings establish temporal error coordination as an important design principle for robot action tokenization, extending beyond the conventional objective of pointwise reconstruction accuracy.

\end{abstract}

\section{Introduction}
\label{sec:introduction}

Precise robotic alignment, such as USB-C insertion, requires successive
small movements. Vision--language--action policies express these commands
through action interfaces \citep{kim2025openvla,pertsch2025fast}.
With low-bit tokens, individually small reconstruction errors can
accumulate when their signs persist. Conversely, errors that cancel
across steps can preserve cumulative commanded motion despite less
accurate individual commands. Pointwise reconstruction misses this distinction.
Figure~\ref{fig:motivation} illustrates cumulative quantization bias.

We study how this temporal organization affects robot action execution.
\textbf{CarryQ} uses the residual left by an earlier token to guide the
next token choice. Based on error feedback
\citep{daubechies2003sigmadelta,karimireddy2019errorfeedback}, it coordinates
existing commands across time within the original interface and budget.
Residuals reset at chunk or query boundaries.

\begin{figure*}[t]
\centering
  \begin{minipage}[t]{0.245\textwidth}%
    \centering\includegraphics[width=\linewidth]{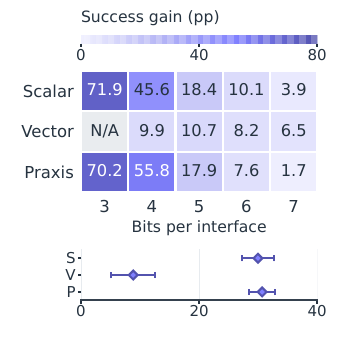}%
    \par\vspace{1pt}{\footnotesize (a) Precision and interface\par}%
  \end{minipage}\hfill
  \begin{minipage}[t]{0.245\textwidth}%
    \centering\includegraphics[width=\linewidth]{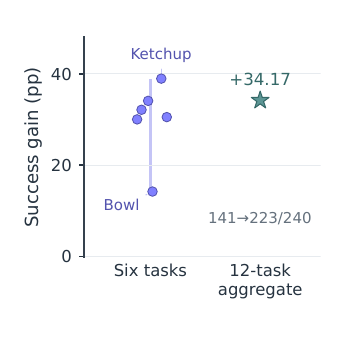}%
    \par\vspace{1pt}{\footnotesize (b) Task breadth\par}%
  \end{minipage}\hfill
  \begin{minipage}[t]{0.245\textwidth}%
    \centering\includegraphics[width=\linewidth]{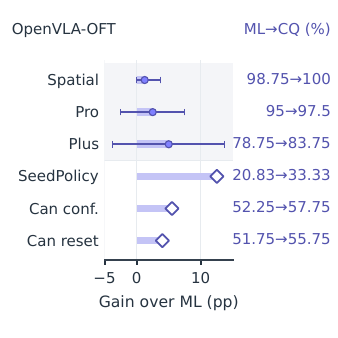}%
    \par\vspace{1pt}{\footnotesize (c) Online effects\par}%
  \end{minipage}\hfill
  \begin{minipage}[t]{0.245\textwidth}%
    \centering\includegraphics[width=\linewidth]{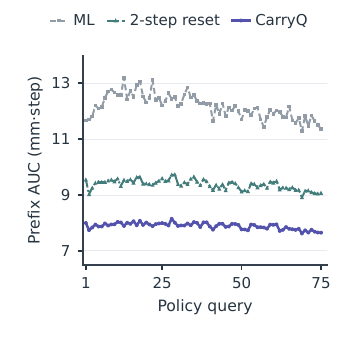}%
    \par\vspace{1pt}{\footnotesize (d) Reset response\par}%
  \end{minipage}
\caption{Execution gains and reset response.
(a) Gains by rate and aggregate 95\% intervals (S: scalar; V: vector;
P: Praxis; vector seed 0; N/A: unevaluated).
(b) Six-task scalar gains and the unconditional 12-task aggregate.
(c) Online gains and success rates; circles include paired 95\% intervals,
diamonds denote aggregate estimates.
(d) Mean prefix error on shared commands, with 75 queries per curve and
four actions per query (Supplementary Section~\ref{app:reset-cohort}).
Execution protocols and sample counts appear in Table~\ref{tab:execution-summary}.}
\label{fig:ef-main}
\end{figure*}

\begin{figure}[t]
  \centering
  \includegraphics[width=\columnwidth]{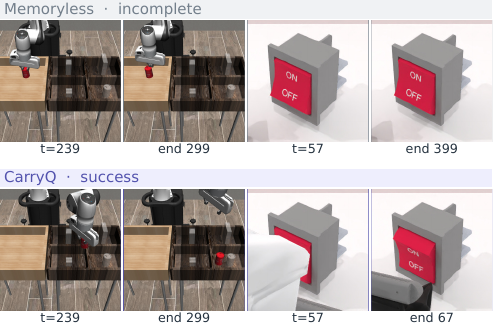}
  \par\vspace{1pt}
  \begin{minipage}[t]{0.49\columnwidth}
    \centering{\footnotesize (a) robomimic Can}
  \end{minipage}\hfill
  \begin{minipage}[t]{0.49\columnwidth}
    \centering{\footnotesize (b) RoboTwin}
  \end{minipage}
  \caption{Paired 4-bit executions with (a) a pretrained Can policy and
  (b) RoboTwin SeedPolicy. CarryQ completes can placement and switch
  turning where memoryless fails. Initial state, camera and crop are
  shared within each pair. Each task shows a matched action index
  followed by terminal states.}
  \label{fig:crossbench-outcomes}
\end{figure}

Controlled replay reveals a reversal: CarryQ improves execution even
when pointwise command error increases. An error decomposition explains
this behavior through stronger cancellation between errors at different
times. Table~\ref{tab:execution-summary} tests the benefit across fixed
interfaces, independently selected tasks and closed-loop policies.
The latter span OpenVLA-OFT on LIBERO, SeedPolicy on RoboTwin and two
robomimic Can sources (Figure~\ref{fig:crossbench-outcomes}), covering
relative motion and absolute joint targets. Fine-insertion simulations
and real-robot charging-connector insertion extend the evaluation to
contact-sensitive completion (Tables~\ref{tab:fine-insertion}
and~\ref{tab:real-robot}).

Our contributions are threefold.
(1) A temporal quantization problem. We formulate cumulative encoding bias
and document it in recorded small, persistent robot commands.
(2) A fixed-interface encoding method. CarryQ transfers residuals across
action steps to promote cancellation while preserving the codebook,
decoder and token budget.
(3) Temporal analysis and execution evidence. Cross-interface replay,
task transfer, online policies and real-robot insertion link temporal
error structure to execution.

\section{Temporal Quantization Bias}
\label{sec:observed-bias}

Recorded robot commands exhibit the persistent bias that motivates
our temporal analysis.
We examine all 120 demonstrations in the 12-task extension, retaining
trajectories independently of their success outcomes.
In the translation coordinates, repeated small commands more often
produce errors with a common sign than larger commands do.
We measure this pattern within same-direction runs of 4--8 steps,
using mean amplitude relative to grid spacing to distinguish small
from larger increments. Action direction and encoding-error sign
are measured separately.

\begin{table}[t]
 \centering
 \caption{Persistent bias in recorded small increments.
 Long runs contain 4--8 same-direction commands; ``one sign'' measures
 strictly same-sign memoryless errors throughout a run.
 ML and CQ columns give endpoint magnitude $|\sum_t e_t|/\Delta_b$.
 Means average runs within demonstrations and then weight tasks equally;
 run counts describe coverage. ML: memoryless; CQ: CarryQ.}
 \label{tab:observed-bias}
 \tablestyle
\begin{tabularx}{\columnwidth}{cYrrrQ}
\toprule
\tablehead
Bits & Amplitude & Runs & One sign (\%) & ML & \textcolor{tableaccent}{CQ} \\
\midrule
4 & Small  & 617   & 49.74 & 1.113 & 0.314 \\
4 & Larger & 5,685 & 5.70  & 0.693 & 0.261 \\
\midrule
6 & Small  & 46    & 26.59 & 0.702 & 0.291 \\
6 & Larger & 6,256 & 3.37  & 0.698 & 0.269 \\
\bottomrule
\end{tabularx}
\end{table}

Table~\ref{tab:observed-bias} shows more frequent same-sign memoryless
error in the small-amplitude strata. CarryQ reduces their endpoint bias
using the same scalar grid. Supplementary
Section~\ref{app:fine-increment} gives coverage and a stricter
sub-bin definition requiring every command to lie below half a grid step.
The effect extends to complete eight-step windows, including zeros and
direction changes (Table~\ref{tab:bias-window-ci}). Thus temporal analysis
examines both the accumulation of persistent rounding and its behavior
across full recorded trajectories.

\section{Related Work}
\label{sec:positioning}

Action representations determine which commands a policy can express.
Coordinate bins, frequency compression and learned vector quantization
offer different choices of this representation
\citep{kim2025openvla,pertsch2025fast,lee2024vqbet,liu2026praxis}.
Chunking and subsequent correction methods organize prediction, selection
and execution over time \citep{zhao2023act,liu2025bid,black2025rtc,wang2026remac}.
Diffusion and flow policies generate continuous chunks
\citep{chi2023diffusionpolicy,lipman2023flowmatching,black2025pi0,shukor2025smolvla};
CorridorVLA introduces spatial anchors and cumulative consistency into flow
training \citep{li2026corridorvla}.
CarryQ studies a complementary decision: how to select existing codewords
across time once the action source and representation have been fixed.
This comparison separates temporal error organization from changes in
the policy, codebook or decoder.

Residual carry is established in sigma-delta modulation and compression
error feedback \citep{daubechies2003sigmadelta,karimireddy2019errorfeedback}.
TCAT studies temporal residual feedback for offline action labels and
fixed-codebook replay \citep{tcat2026}.
CarryQ emphasizes the relationship between local distortion, cross-time
cancellation and execution, examined through recorded bias, matched replay
and online policy-output encoding.
ReAT instead refines an action draft across token stages using a learned,
draft-conditioned decoder and cumulative translation coordinates
\citep{zhan2026reat}. CarryQ retains the decoder and applies feedback across action steps.

\section{CarryQ Encoding}
\label{sec:method}

\begin{figure}[t]
\centering
\includegraphics[width=\columnwidth]{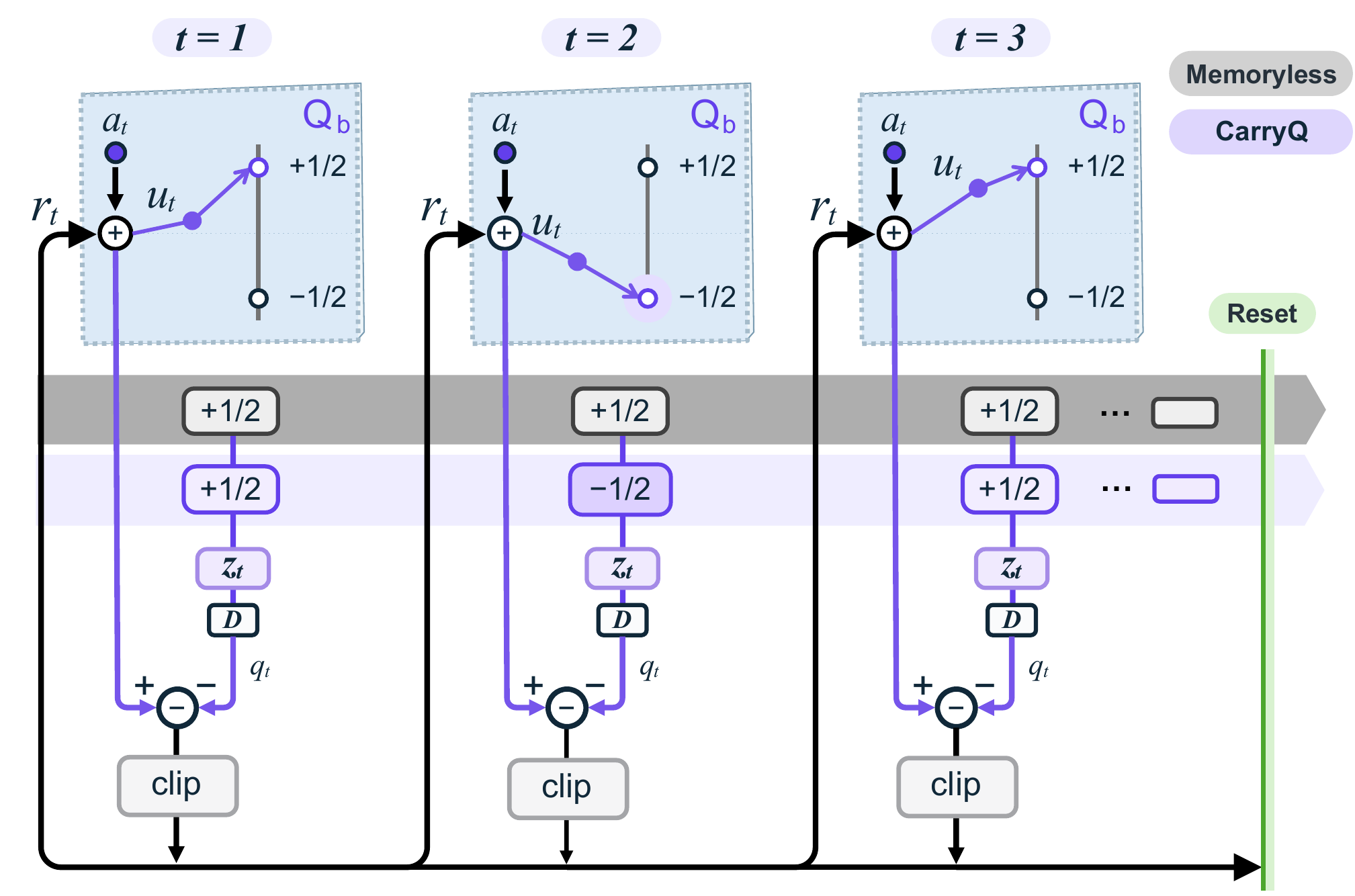}
\caption{CarryQ across action steps. It quantizes $u_t=a_t+r_t$,
decodes $z_t$ to $q_t$, and carries the clipped difference $u_t-q_t$.
Memoryless selects each token independently. Values are in grid units;
feedback resets at window boundaries.}
\label{fig:architecture}
\end{figure}

\subsection{Small-increment bias}

Consider an additive translation coordinate with repeated increment
$a_t=\delta$ over $m$ executed steps. If memoryless encoding maps each
increment to the same grid value $c$, then
\begin{equation}
 e_t=c-\delta,\qquad p_m=m(c-\delta).
 \label{eq:small-increment-bias}
\end{equation}
Local squared error stays at $(c-\delta)^2$, whereas squared cumulative
error grows as $m^2(c-\delta)^2$. Grid spacing, bin location and run
length govern this accumulation. Both tested scalar grids have symmetric,
even-sized vocabularies without an exact zero codeword. A small command
can therefore repeatedly decode to a nonzero value.
Same-direction commands can be consistently over- or under-represented;
their bin locations determine the encoding-error signs.
Section~\ref{sec:observed-bias} documents this bias in recorded commands.

\subsection{Problem formulation}

Let $A=(a_1,\ldots,a_H)$ be an action chunk, $a_t\in\mathbb R^D$,
and $e_t=q_t-a_t$ the error of decoded command $q_t$.
For a coordinate map $S\in\mathbb R^{D'\times D}$, cumulative command
error after $k$ actions is $p_k=\sum_{t=1}^{k}Se_t$.
Persistent same-sign errors reinforce one another, whereas opposing
errors can cancel at the same pointwise MSE. The squared endpoint
separates local magnitude from this temporal interaction:
\begin{equation}
 \left\|\sum_t Se_t\right\|^2
 =\underbrace{\sum_t\|Se_t\|^2}_{L}
 +\underbrace{2\sum_{i<j}\langle Se_i,Se_j\rangle}_{X}.
 \label{eq:error-decomposition}
\end{equation}
Local energy $L$ measures errors independently; the cross-time term $X$
measures their reinforcement or opposition. A negative $X$ can offset
a larger $L$, reducing the squared endpoint $T=L+X$.
For multiaxis commands, the inner products measure directional alignment:
errors pointing in similar directions reinforce deviation, while opposing
directions contribute negative interaction.
For an even-length window, constant errors $e_t=\epsilon$ and
alternating errors $e_t=(-1)^t\epsilon$ both have local energy
$H\epsilon^2$, but their squared endpoints are $H^2\epsilon^2$ and zero.
The cross-time term distinguishes them despite identical pointwise MSE.
Prefix error captures when cancellation occurs: endpoint cancellation can
coexist with large intermediate excursions. Supplementary
Section~\ref{app:decomposition} develops the matrix formulation.

Nearest-codeword selection minimizes current reconstruction error but
omits the cross-products in the cumulative objective. The residual
exposes earlier discrepancies to the next token choice, giving selection
a temporal criterion within the original action set.
In replay, $S$ selects six controller-command coordinates; translation
diagnostics select the first three and normalize by grid spacing.
Supplementary Section~\ref{app:metrics} defines the metrics.

\subsection{Temporal error feedback}
\label{sec:ef-token}

Let $Q_b$ be a $b$-bit quantizer with decoder $\mathcal D$.
Memoryless encoding selects $z_t=Q_b(a_t)$ and decodes
$q_t=\mathcal D(z_t)$ independently. CarryQ instead adds the preceding
encoding residual before selecting the next symbol
(Figure~\ref{fig:architecture}):
\begin{align}
 r_1&=0, &u_t&=a_t+r_t,\nonumber\\
 z_t&=Q_b(u_t), &q_t&=\mathcal D(z_t),\nonumber\\
 r_{t+1}&=\operatorname{clip}
 \left(u_t-q_t,-\frac{\Delta_b}{2},\frac{\Delta_b}{2}\right),
 \label{eq:ef-token}
\end{align}
where $\Delta_b$ is scalar grid spacing and clipping is elementwise.
Residuals use the interface's own decoded values and reset at each chunk
or query. Under-representation leaves a positive residual that raises
the next adjusted input; over-representation gives the opposite correction.
The encoder stores a $D$-dimensional state and adds $O(D)$ arithmetic,
without transmitting residuals or making extra network calls.
The state uses only already-encoded commands and their decoded values,
making the correction available when each token is selected.
Within each shared interface, vocabulary, decoder, token count, execution
window and policy query schedule remain fixed.

Feedback may choose a codeword farther from the current command to
compensate for earlier rounding. When residual clipping and action-range
saturation are inactive, the update gives
\begin{equation}
 q_t-a_t=r_t-r_{t+1},\qquad
 \sum_{t=1}^{H}(q_t-a_t)=-r_{H+1}.
 \label{eq:telescoping}
\end{equation}
Thus the residual left at the window boundary controls its summed error,
whereas persistent memoryless error can grow with window length.
Under these conditions, the residual is the negative accumulated encoding
error. Selecting the codeword nearest $a_t+r_t$ therefore minimizes the
error of the next prefix in the encoded coordinates over the available codewords.
The current command can be represented less accurately while the sequence
becomes more accurate. This connects each token choice to the cumulative
objective using only the error already incurred.
For a coordinate-selection map $S$ and grid spacings $\Delta_{b,d}$ on
selected coordinates $\mathcal I$, every unclipped scalar prefix satisfies
\begin{equation}
 p_k=-Sr_{k+1},\qquad
 \|p_k\|_2\leq\frac12\sqrt{\sum_{d\in\mathcal I}\Delta_{b,d}^{\,2}},
 \quad 1\leq k\leq H.
 \label{eq:prefix-bound}
\end{equation}
The bound follows from $r_1=0$ and the componentwise half-grid residual
bound. It holds at intermediate prefixes as well as the final action,
so compensation need not be deferred to the chunk endpoint. Its scale
depends on grid spacing rather than prefix length.
Resetting removes the residual available to later actions: resetting
every action recovers memoryless encoding, while longer windows permit
compensation. Across reset windows, endpoint residuals contribute to full-sequence
error. Supplementary Section~\ref{app:clipping} reports clipping
under Equation~\ref{eq:ef-token}.

For vector quantization, both methods use shared frozen centroids
$\{c_j\}_{j=1}^M$. Memoryless encoding selects the centroid nearest $a_t$;
CarryQ selects the one nearest
$\bar u_t=\operatorname{clip}(a_t+r_t,-1,1)$ and carries
$\bar u_t-c_{z_t}$. This retains input-range clipping without imposing
a scalar half-grid residual bound. Each method emits one motion token
per action. Native Praxis retains its endpoint-inclusive bins, decoder,
numerical precision and token schedule. Supplementary
Section~\ref{app:token-interfaces} specifies these interfaces.

For continuous-output policies, encoding follows completed prediction
and precedes control. OpenVLA-OFT uses environment commands; the PiPER
comparison shares a frontend that differences absolute joint targets
and a decoder that integrates the decoded increments.
For token-predicting BAR-Smol, CarryQ constructs training labels instead.
Encoding operates along the action sequence after prediction is complete;
it does not modify the internal denoising or flow-sampling iterations of
an action generator. At inference, BAR-Smol decodes its predicted tokens
normally, without reference commands from which to reconstruct a residual.
These placements preserve the source architecture while testing the same
temporal rule in distinct action representations
(Supplementary Section~\ref{app:model-integration}).

\subsection{Execution and metrics}

Budgets are compared within each interface: scalar $b$-bit encoding uses
$b$ bits per coordinate, while a vector token uses $b$ bits for six motion
coordinates jointly. Replay metrics include every recorded action and
partial final chunks. Chunk-summed norm measures remaining error;
maximum prefix norm captures intermediate excursions; rollout-summed
norm measures the full sequence. First-difference MSE measures variation
in decoded commands, rather than physical jerk.

For the 12-task analysis, $L$, $X$ and $T$ are computed per valid chunk
before averaging, preserving cancellation within each chunk.
Mean squared endpoint error differs from the squared mean endpoint norm:
it also captures variation in endpoint magnitudes across chunks.
Complete eight-step windows separately test persistence
at fixed length, including zero commands and direction changes.
Supplementary Section~\ref{app:metrics} defines aggregation and prefix AUC.
Online evaluations pair initial states with independent replanning;
within-query diagnostics share predictions.

\section{Experiments}
\label{sec:experiments}

\begin{figure}[tp]
\centering
\includegraphics[width=\columnwidth]{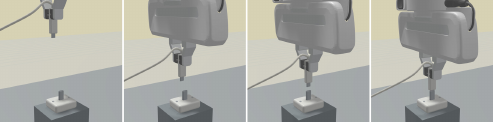}
\par\vspace{1pt}{\footnotesize\textcolor{tableaccent}{(a) USB-C insertion}\par}
\vspace{3pt}
\includegraphics[width=\columnwidth]{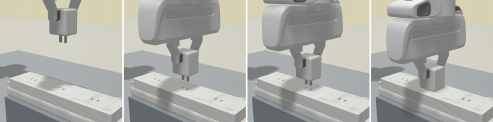}
\par\vspace{1pt}{\footnotesize\textcolor{tableaccent}{(b) Charger insertion}\par}
\par\vspace{4pt}
\centering
  \includegraphics[width=0.326\columnwidth,height=0.326\columnwidth]{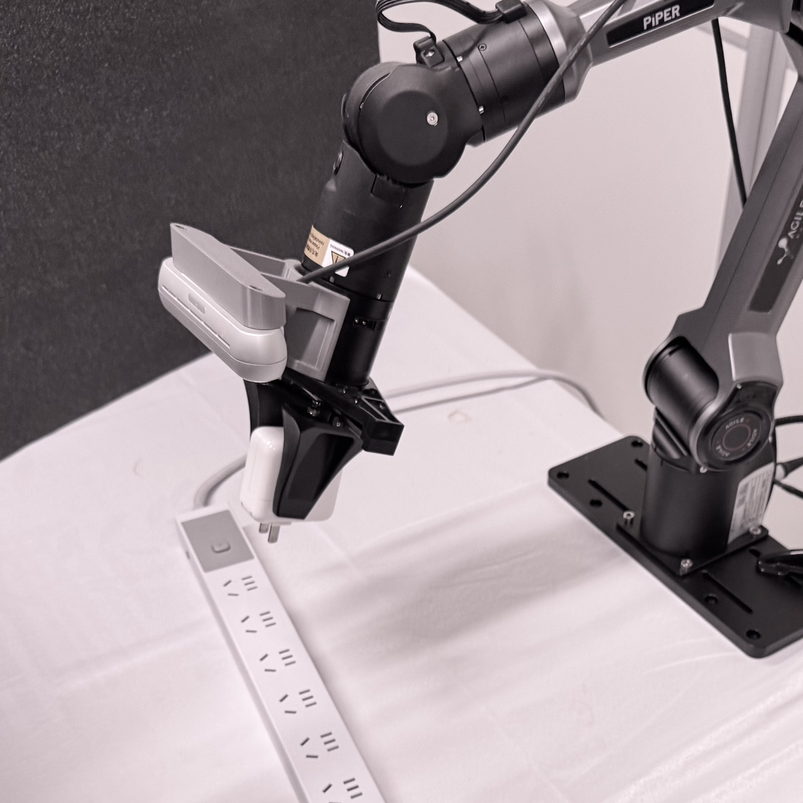}%
  \hfill
  \includegraphics[width=0.326\columnwidth,height=0.326\columnwidth]{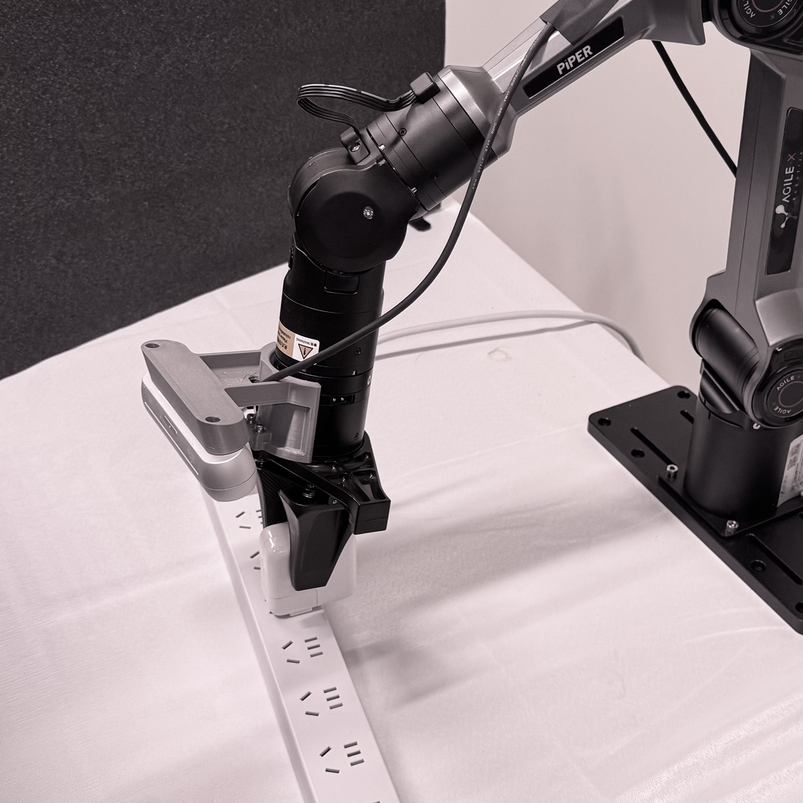}%
  \hfill
  \includegraphics[width=0.326\columnwidth,height=0.326\columnwidth]{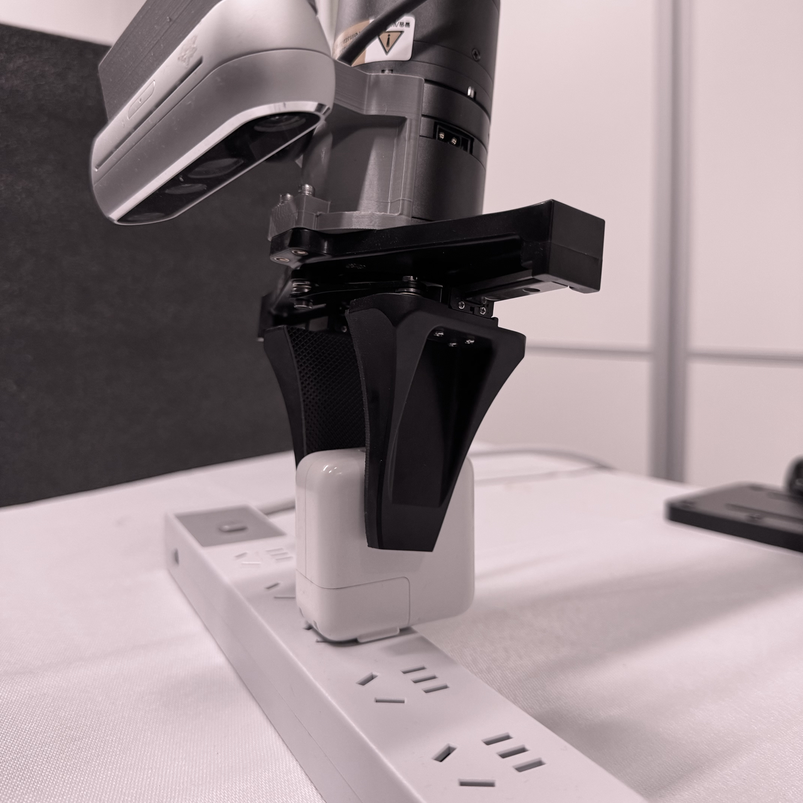}
  \par\vspace{1pt}{\footnotesize\textcolor{tableaccent}{(c) Real-robot insertion}\par}
  \caption{Connector insertion from approach to completion.
  Scripted MuJoCo views show USB-C (a) and charger insertion (b);
  photographs show real-robot insertion into a power strip (c).}
  \label{fig:insertion-task-scenes}
  \label{fig:real-charger-insertion}
\end{figure}

Matched replay isolates temporal encoding, independent tasks test transfer,
and closed-loop policies test its consequences under replanning.
Table~\ref{tab:execution-summary} summarizes simulation outcomes;
Tables~\ref{tab:fine-insertion} and~\ref{tab:real-robot} resolve precision
and stage-wise success in fine insertion.

\subsection{Fine insertion}
\label{sec:fine-insertion}

We evaluate OpenVLA-OFT on USB-C and two-blade charger insertion with a
Franka Panda in MuJoCo (Figure~\ref{fig:insertion-task-scenes}a,b).
The policy predicts $H=8$ actions and executes $K=4$ at 10\,Hz before
replanning. Encoding covers XYZ and three rotation increments; the frozen
policy, action head, preprocessing, decoder and controller are matched.
Each task, method and bit budget has 120 episodes.
S2 denotes complete insertion with electrical connection: full seating
is the connection-success state in simulation, matching the real-robot
completion criterion.

At 8 bits, mean S2 success rises from 15.42\% to 43.33\%
(Table~\ref{tab:fine-insertion}). Both tasks improve at 6 and 4 bits as well.
Higher precision benefits both encoders, while temporal feedback retains
an advantage at every tested budget. At a fixed budget, the decoded action
set is unchanged, so the improvement comes from coordinating available
commands over time. The growing gain at finer precision in these insertion tasks indicates that
feedback and spatial resolution are complementary:
the sequence must progress from alignment to completed contact. Paired 8-bit outcomes contain 35
CarryQ-only versus 3 memoryless-only successes for USB-C, and 38 versus 3
for the charger. Most discordant initial states favor CarryQ in both tasks.
These matched starts link the aggregate improvement to individual
executions that reach full insertion under temporal feedback. Supplementary
Section~\ref{app:fine-insertion} details the protocol.

\begin{table*}[t]
\centering
\caption{Execution success across token interfaces and policy sources.
Gains compare CarryQ with memoryless (ML) at matched budgets, in percentage
points. Six-task replay conditions on continuous-reference success;
the 12-task extension is unconditional. Repeated rates and sources share
initial conditions. The last column gives paired 95\% CIs (pp) or ML-to-CarryQ success counts.
Gain bars share a 0-to-40 pp scale. Policy learning is evaluated separately.}
\label{tab:execution-summary}
\tablestyle
\begin{tabularx}{\textwidth}{YYrQrr}
\toprule
\tablehead
Evaluation & Evaluation scope & ML (\%) & \cqhead\ (\%) & Gain (pp) & Paired evidence \\
\midrule
\tablegroup{6}{A. Fixed-source replay: six tasks, identical source commands and decoders}
Scalar, $H=8$ & 230 demonstrations $\times$ 5 rates & 66.08 & 96.04 & \gainvalue{29.96} & \pairedci{27.27, 32.68} \\
Vector, primary seed 0 & 230 demonstrations $\times$ 4 rates & 70.13 & 78.97 & \gainvalue{8.84} & \pairedci{5.07, 12.59} \\
Praxis tokenizer runtime & 230 demonstrations $\times$ 5 rates & 64.34 & 95.01 & \gainvalue{30.67} & \pairedci{28.46, 32.84} \\
\addlinespace[2pt]
\tablegroup{6}{B. LIBERO task generalization: 12 additional tasks, independently selected}
LIBERO, 4 and 6 bits & 120 demonstrations $\times$ 2 rates & 58.75 & 92.92 & \gainvalue{34.17} & \countchange{141}{223}{240} \\
\addlinespace[2pt]
\tablegroup{6}{C. Online policy-output encoding: matched initial states, per-arm replanning}
OpenVLA-OFT, Spatial & 80 initial states & 98.75 & 100.00 & \gainvalue{1.25} & \pairedci{0.00, 3.75} \\
OpenVLA-OFT, Pro Spatial & 80 initial states & 95.00 & 97.50 & \gainvalue{2.50} & \pairedci{$-$2.50, 7.50} \\
OpenVLA-OFT, Plus Spatial & 80 initial states & 78.75 & 83.75 & \gainvalue{5.00} & \pairedci{$-$3.75, 13.75} \\
OpenVLA-OFT, fine insertion & 2 tasks $\times$ 120 states, 8-bit & 15.42 & 43.33 & \gainvalue{27.92} & \countchange{37}{104}{240} \\
RoboTwin, SeedPolicy & 6 settings $\times$ 20 initial states & 20.83 & 33.33 & \gainvalue{12.50} & \countchange{25}{40}{120} \\
robomimic Can, confirmation & 100 states $\times$ 2 sources $\times$ 2 rates & 52.25 & 57.75 & \gainvalue{5.50} & \countchange{209}{231}{400} \\
robomimic Can, reset & 100 states $\times$ 2 sources $\times$ 2 rates & 51.75 & 55.75 & \gainvalue{4.00} & \countchange{207}{223}{400} \\
\bottomrule
\end{tabularx}
\end{table*}

\begin{table}[t]
\centering
\caption{Fine insertion with OpenVLA-OFT. S2 success (\%), with
120 episodes per task, method and bit budget.}
\label{tab:fine-insertion}
\tablestyle
\begin{tabularx}{\columnwidth}{cYrrQr}
\toprule
\tablehead
Bits & Encoder & USB-C & Charger & Mean & Gain \\
\midrule
8 & Memoryless & 14.17 & 16.67 & 15.42 & \textit{ref.} \\
\rowcolor{tablelight}
8 & \cqhead & 40.83 & 45.83 & \textbf{43.33} & \textcolor{tableaccent}{+27.92} \\
\addlinespace[2pt]
6 & Memoryless & 7.50 & 9.17 & 8.33 & \textit{ref.} \\
\rowcolor{tablefaint}
6 & \cqhead & 26.67 & 30.00 & 28.33 & \textcolor{tableaccent}{+20.00} \\
\addlinespace[2pt]
4 & Memoryless & 2.50 & 3.33 & 2.92 & \textit{ref.} \\
\rowcolor{tablefaint}
4 & \cqhead & 10.00 & 11.67 & 10.83 & \textcolor{tableaccent}{+7.92} \\
\tablenoterow{6}{\columnwidth}{Bits are per motion coordinate. Means and gains use exact counts
and equal task weights. Gain: CarryQ minus same-bit Memoryless (pp).
Primary: 8 bits; precision sensitivity: 6/4 bits.}
\end{tabularx}
\end{table}

\subsection{Replay and transfer}

Six LIBERO-90 tasks \citep{liu2023libero} were selected for quantization
sensitivity and scene diversity using continuous and memoryless replay
before CarryQ evaluation. Demonstrations 10--49 provide held-out commands;
the primary cohort retains 230 conditions with successful continuous replay.
At $H=8$, scalar and native Praxis encoding use 3--7 bits, and vector
encoding uses 4--7 bits. Vector codebooks are fitted on demonstrations
0--9, with seed 0 primary and two further seeds testing stability.
Encoder pairs share source actions, initial and model states, codewords,
decoder, budget and window. Success weights tasks equally after averaging
rates within demonstrations. Paired 95\% intervals resample whole
demonstrations within task, retaining rates and encoders together
(Supplementary Sections~\ref{app:replay-tasks} and~\ref{app:bootstrap}).

\begin{figure*}[t]
  \centering
  \includegraphics[width=\textwidth]{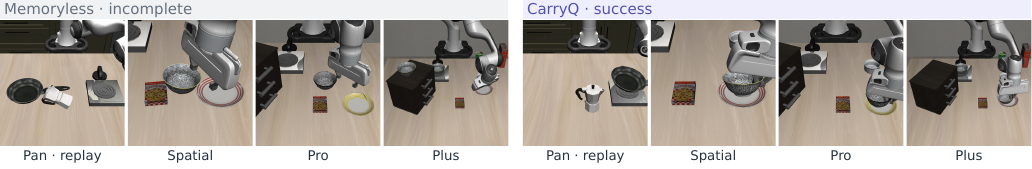}
  \caption{LIBERO outcomes at 4 bits. Left: memoryless; right: CarryQ,
  with the same four tasks in each group. Pan uses recorded-source replay;
  Spatial, Pro and Plus use closed-loop OpenVLA-OFT. Matched terminal
  views share initial state, camera and crop.}
  \label{fig:replay-outcomes}
\end{figure*}

CarryQ improves execution across all three interfaces
(Table~\ref{tab:execution-summary}). Scalar success rises from 66.08\% to 96.04\%, and all six tasks improve
(Figure~\ref{fig:ef-main}b). Vector gains remain positive at every tested
rate and fitting seed; native Praxis gains span all tasks and rates.
Because each vector comparison freezes its centroids, the gain reflects
how existing codewords are selected across time. The differing grids
and vector partitions test this rule without changing decoded action
capacity. Figure~\ref{fig:replay-outcomes} shows a replay outcome; Supplementary
Section~\ref{app:replay-sequences} extends all three interfaces to longer sequences.

Native Praxis also improves execution despite higher pointwise command
MSE (Supplementary Section~\ref{app:token-interfaces}). Thus the reversal
between reconstruction and execution extends beyond the prescribed scalar
grid. Figure~\ref{fig:ef-main}(a) shows the largest scalar and Praxis gains
at coarse rates; the gap narrows as memoryless success approaches its
ceiling. Temporal coordination complements the resolution provided by
additional codewords.

A prespecified hash-based procedure selects 12 additional tasks
independently of CarryQ outcomes. All 120 demonstration conditions are
evaluated at two budgets, regardless of reference success.
Unconditional task-balanced success rises from 58.75\% to 92.92\%
(Table~\ref{tab:execution-summary}; Figure~\ref{fig:ef-main}b).
The independent cohort extends the benefit beyond the initial sensitivity
screen, with equal task weights and all reference outcomes retained.
Supplementary
Section~\ref{app:task-transfer} lists the tasks and sampling design.

\subsection{Temporal error structure}

\begin{figure}[t]
\centering
  \begin{minipage}[t]{0.485\columnwidth}%
    \centering\includegraphics[width=\linewidth]{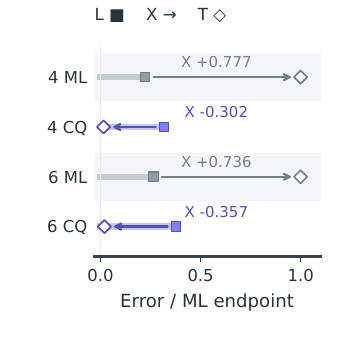}%
    \par\vspace{1pt}{\footnotesize (a) Error decomposition\par}%
  \end{minipage}\hfill
  \begin{minipage}[t]{0.485\columnwidth}%
    \centering\includegraphics[width=\linewidth]{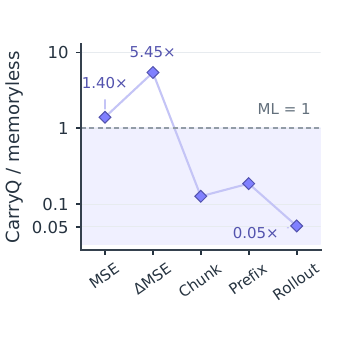}%
    \par\vspace{1pt}{\footnotesize (b) Error profile\par}%
  \end{minipage}
\caption{Cross-time cancellation offsets greater local distortion.
(a) The 12-task $L+X=T$ decomposition, normalized by each rate's memoryless $T$;
squares, arrows and diamonds denote $L$, $X$ and $T$.
(b) Six-task scalar error ratios on a log scale, with paired 95\% intervals;
$\Delta$MSE measures decoded-command first differences.
Supplementary Sections~\ref{app:bootstrap} and~\ref{app:decomposition} give definitions.}
\label{fig:decomposition}
\end{figure}

\begin{figure}[t]
\centering
  \begin{minipage}[t]{0.485\columnwidth}%
    \centering\includegraphics[width=\linewidth]{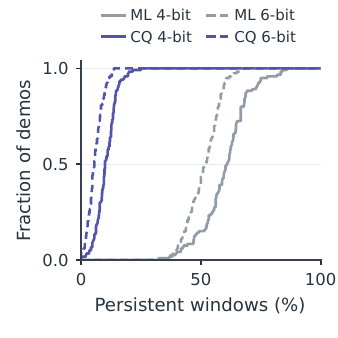}%
    \par\vspace{1pt}{\footnotesize (a) Persistence distribution\par}%
  \end{minipage}\hfill
  \begin{minipage}[t]{0.485\columnwidth}%
    \centering\includegraphics[width=\linewidth]{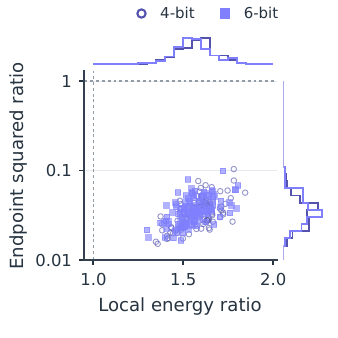}%
    \par\vspace{1pt}{\footnotesize (b) Distortion reversal\par}%
  \end{minipage}
\caption{Complete-window evidence across 120 demonstrations.
(a) Error-sign persistence distributions at 4 bits (solid) and 6 bits
(dashed). (b) All 240 demonstration--rate pairs show greater local energy
and lower squared endpoint error under CarryQ; margins show distributions.
The two rates share demonstrations (Supplementary Section~\ref{app:complete-windows}).}
\label{fig:temporal-bias}
\end{figure}

CarryQ changes the relationship between local distortion and cumulative
command accuracy. In six-task scalar replay, pointwise MSE rises by
40.0\%, while mean chunk-summed error norm falls by 87.31\%
(Table~\ref{tab:ef-mechanism}). Lower maximum prefix and rollout-summed
norms show that compensation reduces discrepancies at intermediate command
prefixes and across complete sequences (Figure~\ref{fig:decomposition}b).
The rollout-summed error combines the residual discrepancies from successive
windows. Its magnitude depends on how they align, whereas the mean chunk
norm measures each window separately. Reductions in both summaries show
that compensation improves recorded commands at the window and rollout scales.
Feedback may cross a codeword boundary to offset prior rounding, increasing
first-difference MSE while reducing cumulative discrepancy.
The additional variation can therefore serve a temporal purpose:
correcting accumulated error in the decoded command sequence.
These measurements include all recorded actions, including clipped updates;
Supplementary Section~\ref{app:clipping} quantifies their incidence.

\begin{table}[t]
 \centering
 \caption{Local distortion and cumulative command error for scalar $H=8$
 encoding in the first six controller-command coordinates.
 First-difference MSE measures decoded-command variation.
 Arrows indicate change relative to memoryless; Supplementary Section~\ref{app:bootstrap}
 reports demonstration-cluster intervals.}
 \label{tab:ef-mechanism}
 \tablestyle
\begin{tabularx}{\columnwidth}{YrQr}
\toprule
\tablehead
Measure & Baseline & \cqhead & Change \\
\midrule
\tablegroup{4}{Local distortion and command variation}
Pointwise command MSE & 0.002309 & 0.003233 & $\uparrow$\,40.0\% \\
Decoded first-difference MSE & 0.002163 & 0.011776 & 5.45$\times$ \\
\addlinespace[2pt]
\tablegroup{4}{Cumulative command error}
Mean chunk-summed norm & 0.5110 & \textbf{0.0648} & $\downarrow$\,87.31\% \\
Maximum prefix norm & 0.5131 & \textbf{0.0956} & $\downarrow$\,81.36\% \\
Rollout-summed norm & 5.923 & \textbf{0.305} & $\downarrow$\,94.85\% \\
\bottomrule
\end{tabularx}
\end{table}

The decomposition explains the reversal (Figure~\ref{fig:decomposition}a).
On the 12-task traces, $X$ changes from positive under memoryless encoding
to negative under CarryQ at both precisions. Errors that previously
reinforced one another now cancel more strongly across steps.
This cancellation offsets the larger local energy $L$, reducing squared
endpoint error $T=L+X$ by about 98\%.
Ranking encoders only by local distortion would therefore reverse their
ranking by cumulative command accuracy. Supplementary
Section~\ref{app:decomposition} reports the component values and
prefix summaries.

\begin{table}[t]
\centering
\caption{Complete-window temporal error on 120 demonstrations.}
\label{tab:bias-window-ci}
\tablestyle
\begin{tabularx}{\columnwidth}{YrrQr}
\toprule
\tablehead
Measure & Bits & ML & \cqhead & Reduction {\tablesubfont\color{tableaccent}[95\% CI]} \\
\midrule
Persistence (\%) & 4 & 60.25 & 10.50 & \inlineci{49.75}{48.59, 50.90} \\
 & 6 & 51.26 & 5.85 & \inlineci{45.41}{44.20, 46.65} \\
\midrule
Endpoint / $\Delta_b$ & 4 & 1.157 & 0.237 & \inlineci{0.921}{0.894, 0.949} \\
 & 6 & 1.150 & 0.242 & \inlineci{0.908}{0.881, 0.934} \\
\tablenoterow{5}{\columnwidth}{Persistence: windows with $\geq4$ consecutive same-sign errors.
Endpoint: mean magnitude. Reductions are ML minus CarryQ, in pp or grid
units; 95\% intervals cluster by demonstration (Supplementary Section~\ref{app:complete-windows}).}
\end{tabularx}
\end{table}

Complete windows test this structure beyond selected command strata.
Across 120 demonstrations, error-sign persistence decreases at both
budgets (Figure~\ref{fig:temporal-bias}a). All 240 demonstration--rate
pairs have larger local energy and smaller squared endpoint error
(Figure~\ref{fig:temporal-bias}b), so the reversal is not confined to aggregate
means. The rates share demonstrations; Table~\ref{tab:bias-window-ci}
reports paired intervals. Supplementary
Section~\ref{app:bias-outcomes} examines associations with task benefit.

\subsection{Real-robot insertion}
\label{sec:real-robot}

We evaluate Charging Connector Insertion using $\pi_{0.5}$
\citep{physicalintelligence2025pi05} on PiPER with external and wrist
RealSense D455 views, at 10\,Hz and $K=4$ actions per query.
Both encoders share joint-target differencing and an integrating decoder;
residuals reset at each query and gripper commands bypass encoding.
Each of seven configurations has 50 trials. Alignment (S1) records an
insertable pose; S2 records complete insertion with electrical connection,
verified by a meter. Both stage rates use all 50 trials.
Figure~\ref{fig:real-charger-insertion}(c) shows this sequence.

\begin{table}[t]
\centering
\caption{Charging Connector Insertion with $\pi_{0.5}$ on PiPER
at 10\,Hz and $K=4$, with 50 trials per configuration.}
\label{tab:real-robot}
\tablestyle
\begin{tabularx}{\columnwidth}{YcrQrr}
\toprule
\tablehead
Encoder & Bits & S1 & S2 & $\Delta$S1 & $\Delta$S2 \\
\midrule
Continuous & Full & 19 (38) & 9 (18) & \multicolumn{2}{c}{\textit{source}} \\
\tablegroup{6}{Joint-increment encoding}
Memoryless & 4 & 10 (20) & 3 (6) & \multicolumn{2}{c}{\textit{baseline}} \\
\rowcolor{tablefaint}
\cqhead & 4 & 27 (54) & \textbf{18 (36)} & \textcolor{tableaccent}{+34} & \textcolor{tableaccent}{+30} \\
Memoryless & 6 & 17 (34) & 8 (16) & \multicolumn{2}{c}{\textit{baseline}} \\
\rowcolor{tablefaint}
\cqhead & 6 & 32 (64) & \textbf{21 (42)} & \textcolor{tableaccent}{+30} & \textcolor{tableaccent}{+26} \\
\tablegroup{6}{Absolute-target controls}
Memoryless & 4 & 8 (16) & 2 (4) & \multicolumn{2}{c}{\textit{control}} \\
Memoryless & 6 & 14 (28) & 6 (12) & \multicolumn{2}{c}{\textit{control}} \\
\tablenoterow{6}{\columnwidth}{S1: alignment; S2: complete insertion with electrical connection.
Stage cells show successes (\%) among 50 trials.
$\Delta$: gain over same-bit Memoryless (pp).
Absolute-target rows are representation controls.}
\end{tabularx}
\end{table}

CarryQ improves both stages at each tested precision
(Table~\ref{tab:real-robot}). Final success rises from 6\% to 36\% at 4 bits
and from 16\% to 42\% at 6 bits, reaching completed electrical connection
beyond alignment. Both rates also exceed the continuous policy's 18\%
completion. The primary CarryQ--memoryless comparison shares the incremental
representation, reconstruction and controller, isolating temporal encoding.
Together, S1 and S2 show improvements from reaching an insertable pose
through completing the contact sequence.
Supplementary Section~\ref{app:real-robot} gives the reconstruction,
control path and evaluation protocol.

\subsection{Closed-loop policies}

We also evaluate OpenVLA-OFT on LIBERO, SeedPolicy on RoboTwin, and two
robomimic Can sources, holding each checkpoint fixed within comparisons.
Arms start from paired initial states, then observe their own executions
and replan. Shared-command replay isolates the encoding transformation;
closed-loop evaluation tests how its consequences feed back into subsequent
predictions. Together, these settings connect temporal encoding to execution.
These interfaces span relative Cartesian increments and absolute joint
targets. Their common intervention is temporal codeword selection
within each controller interface, retaining the policy's action
representation and generation process (Supplementary
Sections~\ref{app:model-integration} and~\ref{app:closed-loop}).

OpenVLA-OFT has positive mean differences in Spatial, Pro Spatial and
Plus Spatial, with 80 initial states per environment
(Table~\ref{tab:execution-summary}; Figure~\ref{fig:ef-main}c).
Standard spatial tasks are near the memoryless ceiling; object,
instruction and observation or layout variations provide different
execution conditions. Figure~\ref{fig:replay-outcomes} compares terminal
outcomes; Supplementary Section~\ref{app:oft-evaluation} gives sequences and prediction checks.

RoboTwin changes both the environment and action semantics. SeedPolicy
predicts absolute joint targets for cup placement, bowl stacking and switch
turning in six clean or randomized settings, each with 20 initial states.
CarryQ gains 12.50 percentage points while retaining TOPP and drive-target
control. Figure~\ref{fig:crossbench-outcomes}(b) illustrates switch turning;
Supplementary Section~\ref{app:seedpolicy} gives setting-level controls.

Can tests translation-only encoding under low-dimensional observations,
using pretrained and publicly DSRL-adapted sources
\citep{wagenmaker2025steering}. Rotation and gripper commands pass through
unchanged. Each source is evaluated at 4 and 6 bits on 100 shared initial
states, yielding 400 executions per encoder.
CarryQ improves confirmation-cohort success by 5.50 points and exceeds
three-seed random rounding in aggregate success, 57.75\% versus 51.08\%.
Random rounding varies token choices through sampling, whereas CarryQ
conditions each choice on the error already incurred. The comparison
therefore tests the value of directing that variation toward compensation
within each execution window. Repeating random rounding across seeds
and retaining the same source policies separates the feedback rule from
a particular sampled token sequence.
Figure~\ref{fig:crossbench-outcomes}(a) shows can placement; Supplementary
Section~\ref{app:can-confirmation} details both source policies.

\subsection{Reset and robustness}

\begin{table}[t]
\centering
\caption{Codebook robustness, resets and online error.}
\label{tab:robustness-consistency}
\tablestyle
\begin{tabularx}{\columnwidth}{Yr}
\toprule
\tablehead
Setting or measure & Result \\
\midrule
\tablegroup{2}{Codebook robustness: success gain (pp)}
Seeds 0, 1, 2 & +8.84, +9.04, +8.11 \\
\tablegroup{2}{Independent Can reset: success (\%)}
Memoryless & \inlinecount{51.75}{207/400} \\
2-step reset & \inlinecount{46.25}{185/400} \\
CarryQ (4-step) & \inlinecount{\textbf{55.75}}{223/400} \\
\tablegroup{2}{Online command consistency}
Can prefix-AUC reduction: Native & 33.99\%, 32.71\% \\
Can prefix-AUC reduction: DSRL & 37.31\%, 32.71\% \\
Can cells with lower prefix AUC & 400/400 \\
SeedPolicy queries with lower error & 30,977/30,977 \\
\tablenoterow{2}{\columnwidth}{Can reductions list 4- and 6-bit results.
Each Can cohort uses 100 states across two sources and two rates;
cohorts remain separate. SeedPolicy measures predicted chunk-sum error,
including unexecuted tails; queries are not independent success trials.}
\end{tabularx}
\end{table}

The temporal benefit also appears in command diagnostics
(Table~\ref{tab:robustness-consistency}). Can lowers translation-command
prefix AUC despite larger local energy; SeedPolicy lowers predicted
chunk-sum error. Can measures four issued actions, whereas SeedPolicy
measures complete predicted joint-target chunks, including unexecuted
tails. The diagnostics thus retain the coordinates and windows of each
interface rather than pooling unlike error magnitudes.

An independent Can cohort uses 100 new initial states with the same two
sources and rates. It holds four executed actions fixed while resetting
residuals every two actions or after the query.
Four-step CarryQ gains 4.00 points over memoryless encoding; the two-step
variant lowers aggregate success (Table~\ref{tab:robustness-consistency}).
The latter discards the first pair's residual before encoding the second
pair, isolating residual lifetime at unchanged query frequency.
On shared source commands, CarryQ has lower mean prefix AUC than two-step
reset at all 75 queries (Figure~\ref{fig:ef-main}d; Supplementary
Section~\ref{app:reset-cohort}).
Learning CarryQ targets with BAR-Smol, a Praxis autoregressive policy
using SmolVLM \citep{liu2026praxis,marafioti2025smolvlm}, is evaluated
separately in Supplementary Section~\ref{app:encoded-target-learning}.

\section{Conclusion}

Action tokenization depends on error interactions across time as well as
individual magnitudes. CarryQ coordinates existing token
choices through residual feedback, improving execution across replay,
closed-loop policies and connector insertion. The decomposition $T=L+X$
explains how stronger cross-time cancellation reduces cumulative command
error despite greater local distortion. Local fidelity, temporal structure and task success together capture
benefits missed by pointwise MSE.

\label{carryq:body-end}

{\small
\setlength{\bibsep}{0pt}
\setlength{\baselineskip}{10.5pt}

}

\clearpage
\appendix
\twocolumn[\centering\Large
  \textbf{\thetitle}\\
  \vspace{0.5em}Appendix\\
  \vspace{1.0em}
]
\setcounter{figure}{0}
\setcounter{table}{0}
\setcounter{equation}{0}
\renewcommand{\thefigure}{S\arabic{figure}}
\renewcommand{\thetable}{S\arabic{table}}
\renewcommand{\theequation}{S\arabic{equation}}
\renewcommand{\theHfigure}{appendix.\arabic{figure}}
\renewcommand{\theHtable}{appendix.\arabic{table}}
\renewcommand{\theHequation}{appendix.\arabic{equation}}
\setcounter{dbltopnumber}{1}
\section{Interfaces and Notation}
\label{app:model-integration}

Section~\ref{app:insertion-evaluation} covers simulated and real insertion;
Section~\ref{app:reproducibility} gives replay and task-transfer protocols;
Section~\ref{app:fine-increment} develops temporal diagnostics;
Sections~\ref{app:closed-loop} and~\ref{app:reset-learning} cover closed-loop
controls, resets and learned targets. Figures and tables are indexed
within these topics.

\subsection{Shared encoding stage}
\label{app:integration-common}

CarryQ encodes commands from completed policy predictions, recorded
demonstrations or training targets. For a predicted chunk
$A=(a_1,\ldots,a_H)$ and current state $s_0$, let $f$ be the common
coordinate frontend and $g$ its reconstruction:
\begin{equation}
 U=f(A_{1:K};s_0),\qquad
 \widehat A_{1:K}=g\bigl(\mathcal D(Z);s_0\bigr).
 \label{eq:integration-interface}
\end{equation}
Memoryless encoding and CarryQ differ in selecting $Z$ from $U$;
the frontend, reconstruction, codewords, bounds and control path are
matched. $H$ is the prediction horizon and $K$ the selected execution
window. For direct encoding, $f$ and $g$ preserve source coordinates;
PiPER instead uses differencing and integration.
Table~\ref{tab:model-integration} summarizes the placements.
In frozen-policy comparisons, the visual encoder, language conditioning,
action generator and query schedule stay fixed. Each arm then observes
its own execution and replans. Residuals reset at every query; unexecuted
actions do not carry into the next query.

\begin{table}[htbp]
\centering
\caption{CarryQ interfaces. Windows are encoding and reset units;
execution may end before all predicted actions are used.}
\label{tab:model-integration}
\tablestyle
\begin{tabularx}{\columnwidth}{>{\raggedright\arraybackslash}p{0.20\columnwidth}YY}
\toprule
\tablehead
Source & Encoding & Reconstruction \\
\midrule
$\pi_{0.5}$, PiPER & Six joint increments; $K=4$ & Integrate from joint anchor; gripper bypass \\
OpenVLA-OFT, LIBERO & Six motion coordinates; $H=8$ & Environment commands; gripper unchanged \\
OpenVLA-OFT, insertion & Six motion increments; $H=8$, $K=4$ & Same decoder and controller \\
SeedPolicy & 12 joint targets; $H=6$ & TOPP and drive targets; two grippers bypass \\
Can, two sources & XYZ; up to $K=4$ & Original action conversion; other channels bypass \\
Praxis replay & Native motion tokens; $H=8$ & Native decoder, schedule and gripper path \\
BAR-Smol & Training tokens; $H=8$ & Original decoder for predicted tokens \\
\tablenoterow{3}{\columnwidth}{Policy-output encoders reset at each query;
replay and training encoders reset at chunk boundaries.}
\end{tabularx}
\end{table}

\subsection{Token interfaces}
\label{app:token-interfaces}

Scalar replay uses midpoint bins on the first six normalized command
coordinates, with spacing $2/2^b$. Vector replay uses task-specific
K-means centroids, frozen after fitting; one token represents all six
motion coordinates. Both bypass the gripper and reset every $H=8$
actions. Vector success gains are +8.84, +9.04 and +8.11 percentage
points for fitting seeds 0, 1 and 2.

Native Praxis retains endpoint-inclusive centers with spacing
$2/(2^b-1)$, float32 arithmetic, the flattened token schedule, vocabulary,
decoder and native gripper treatment. Task-balanced pointwise MSE rises
from 0.002969 to 0.004162 alongside its execution gain. This native
comparison is distinct from the endpoint-grid clipping idealization in
Section~\ref{app:clipping}. Both scalar grids are symmetric and have
an even number of codewords, hence no exact zero codeword.

\begin{figure*}[t]
\centering
\includegraphics[width=\textwidth]{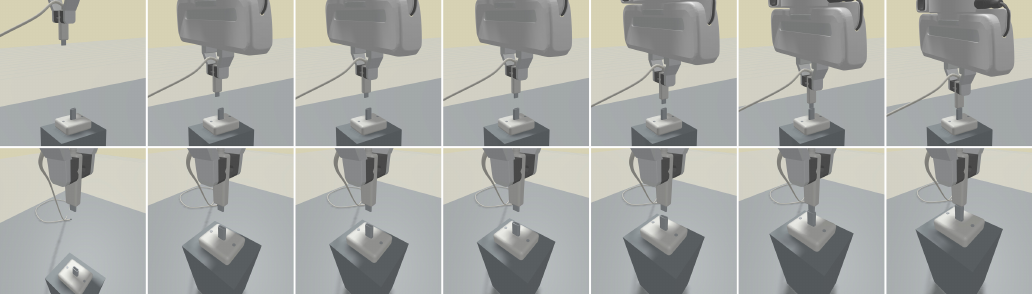}
\caption{USB-C insertion task illustration. External (top) and wrist (bottom) views share seven saved simulation states.}
\label{fig:app-usb-task-scene}
\end{figure*}

\begin{figure*}[t]
\centering
\includegraphics[width=\textwidth]{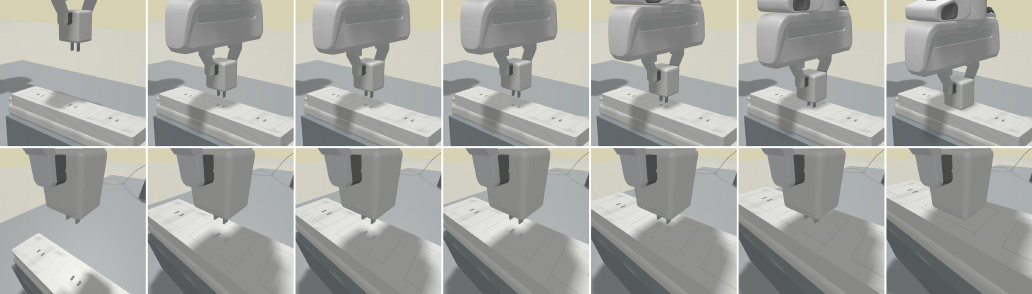}
\caption{Charger insertion into a power strip. External (top) and wrist (bottom) views share seven saved simulation states.}
\label{fig:app-charger-task-scene}
\end{figure*}

\subsection{Policy coordinates}
\label{app:other-integrations}

OpenVLA-OFT on LIBERO encodes the completed eight-action prediction after
conversion to environment commands. Six motion coordinates use 16
endpoint-inclusive centers with fixed first- and 99th-percentile
action-statistics bounds. The categorical gripper is unchanged.
CarryQ uses a half-step residual bound and resets at each query.
The separate fine-insertion setting uses 8/6/4-bit encoding and executes
four of eight predicted actions (Section~\ref{app:fine-insertion}).

SeedPolicy supplies six actions with 14 channels. Encoding follows
unnormalization and covers 12 absolute arm-joint targets using 16
endpoint-inclusive centers with fixed action-statistics minima and maxima.
The two grippers bypass encoding. Decoded targets retain TOPP and drive-target control.
PiPER instead integrates increments.

Both Can sources encode translation only, before inverse action
normalization, using 4- or 6-bit endpoint grids on the selected prefix
of up to four actions. Rotation and gripper commands pass through.
Controller scaling and saturation remain in place; diagnostics are
computed after conversion, in the physical translation scale.
For $\pi_{0.5}$, encoding follows the flow-based action expert and output
transforms, where the client receives absolute joint targets. Its shared
incremental interface is specified in Section~\ref{app:real-robot}.
BAR-Smol instead learns CarryQ-encoded targets; predicted tokens use
the original decoder (Section~\ref{app:encoded-target-learning}).

\section{Fine Insertion}
\label{app:insertion-evaluation}
\label{app:insertion-task-scenes}

\subsection{Simulation protocol}
\label{app:fine-insertion}

Both insertion tasks use OpenVLA-OFT on a seven-DoF Franka Panda in MuJoCo.
The source policy predicts eight actions and executes four before the
next query, at a 10\,Hz control frequency. The six encoded coordinates
are XYZ increments and three rotation increments, with the reported
number of bits per coordinate.
The policy checkpoint, action head, preprocessing, decoder and low-level
controller are shared between memoryless encoding and CarryQ.
We evaluate 8-bit encoding as the primary setting and 6/4 bits as a
precision-sensitivity study, with 120 episodes per task, method and rate.
Each group comprises six batches of 20 distinct initial states.
The matrix has 1,440 episodes.

S2 denotes complete connector insertion with electrical connection.
Complete seating is the connection-success condition in simulation;
the real-robot experiment uses the same completion criterion with a
measurement device verifying electrical connection.
Each episode ends after completion or at 600 control steps (60\,s at 10\,Hz).
Task rates use all 120 episodes as the denominator. Mean success weights
the two tasks equally, so it also equals pooled successes divided by 240
at each method and precision.

\begin{table}[tbp]
\centering
\caption{Fine-insertion S2 successes and rates. Each task, encoder
and bit budget has 120 episodes.}
\label{tab:app-fine-insertion-counts}
\tablestyle
\begin{tabularx}{\columnwidth}{cYrrQ}
\toprule
\tablehead
Bits & Encoder & USB-C & Charger & Mean (\%) \\
\midrule
8 & Memoryless & 17/120 (14.17) & 20/120 (16.67) & 15.42 \\
\rowcolor{tablefaint}
8 & \cqhead & 49/120 (40.83) & 55/120 (45.83) & 43.33 \\
\addlinespace[2pt]
6 & Memoryless & 9/120 (7.50) & 11/120 (9.17) & 8.33 \\
\rowcolor{tablefaint}
6 & \cqhead & 32/120 (26.67) & 36/120 (30.00) & 28.33 \\
\addlinespace[2pt]
4 & Memoryless & 3/120 (2.50) & 4/120 (3.33) & 2.92 \\
\rowcolor{tablefaint}
4 & \cqhead & 12/120 (10.00) & 14/120 (11.67) & 10.83 \\
\tablenoterow{5}{\columnwidth}{Task cells show successes/120 (\%).
Mean weights the two tasks equally; gain values are in the main table.}
\end{tabularx}
\end{table}

Figures~\ref{fig:app-usb-task-scene} and~\ref{fig:app-charger-task-scene}
show scripted MuJoCo sequences from approach
through alignment to seating. Each column pairs external and wrist views
of the same saved state. External cameras remain fixed; wrist cameras
move rigidly with the hand. These illustrations describe task geometry;
Tables~\ref{tab:app-fine-insertion-counts} and~\ref{tab:app-fine-insertion-pairs}
report the policy evaluations.

\subsection{Paired outcomes}
\label{app:fine-insertion-pairs}

Table~\ref{tab:app-fine-insertion-pairs} preserves the paired S2 outcomes
at 8 bits. USB-C has 35 CarryQ-only successes and 3 memoryless-only
successes; the charger has 38 and 3. Their net gains are 32 and 35
successes out of 120, respectively. Across tasks, 37 memoryless successes
and 104 CarryQ successes yield a mean gain of 27.92 percentage points.
The corresponding aggregate gains at 6 and 4 bits are 20.00 and 7.92 points.
The 8-bit mean rate is 2.81 times Memoryless.

\begin{table}[tbp]
\centering
\caption{Paired 8-bit S2 outcomes on shared initial states.}
\label{tab:app-fine-insertion-pairs}
\tablestyle
\begin{tabularx}{\columnwidth}{Yrrr}
\toprule
\tablehead
Outcome & USB-C & Charger & Total \\
\midrule
Both succeed & 14 & 17 & 31 \\
\rowcolor{tablefaint}
CarryQ only & 35 & 38 & 73 \\
Memoryless only & 3 & 3 & 6 \\
Both fail & 68 & 62 & 130 \\
\midrule
Initial-state pairs & 120 & 120 & 240 \\
\tablenoterow{4}{\columnwidth}{Each task uses 120 initial-state pairs; totals sum both tasks.}
\end{tabularx}
\end{table}

\begin{figure*}[t]
  \centering
  \includegraphics[width=\textwidth]{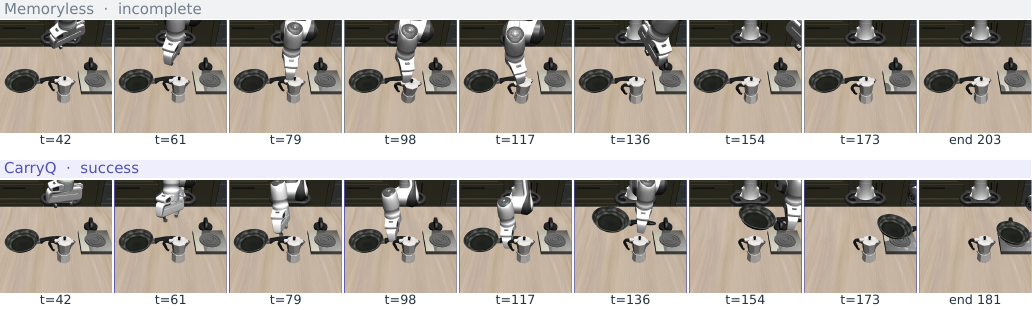}
  \caption{Scalar action encoding on the frying-pan task at 4 bits.
  Demonstration 22 retains the median-completion selection in
  Section~\ref{app:replay-sequences}. The memoryless trajectory misses the
  stove, while CarryQ with $H=8$ reaches the task criterion at action 181.
  Ordinary columns match action indices; the last column shows each arm's
  terminal state. Both rows share recorded source commands, initial state,
  camera and crop.}
  \label{fig:app-e3-sequence}
\end{figure*}

\begin{figure*}[t]
  \centering
  \begin{minipage}[t]{0.477\textwidth}
    \noindent{\small a}\par\vspace{1pt}
    \includegraphics[width=\linewidth]{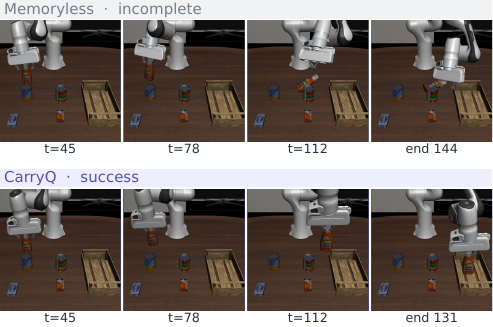}
  \end{minipage}\hfill
  \begin{minipage}[t]{0.477\textwidth}
    \noindent{\small b}\par\vspace{1pt}
    \includegraphics[width=\linewidth]{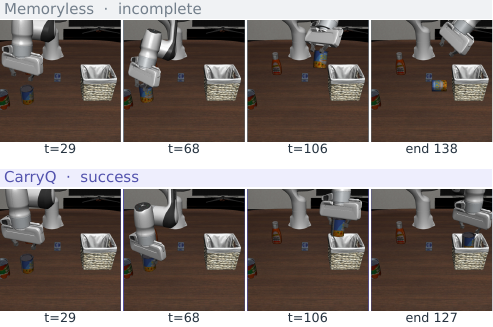}
  \end{minipage}
  \caption{Native Praxis action encoding at 4 bits.
  Ketchup demonstration 10 (a) and soup demonstration 10 (b) retain the
  selected anchors described in Section~\ref{app:replay-sequences}.
  CarryQ reaches the task criterion at actions 131 and 127, respectively;
  memoryless encoding fails in both cases. Each pair shares recorded source
  commands, initial state and native decoder. Ordinary columns match action
  indices; each final column shows the two actual terminal states.}
  \label{fig:app-e5-sequences}
\end{figure*}

\begin{figure*}[t]
  \centering
  \includegraphics[width=\textwidth]{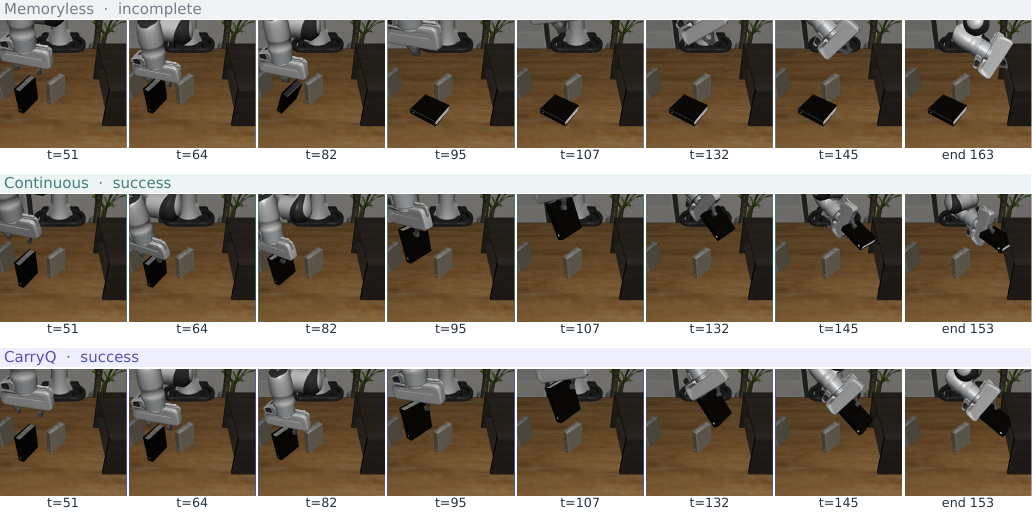}
  \caption{Vector-codebook encoding on book demonstration 36 at 4 bits,
  with fitting seed 0. Rows show memoryless encoding, the unquantized
  continuous source and CarryQ. The memoryless trajectory drops the book
  onto the table; the continuous source and CarryQ reach the shelf goal.
  All rows use the same initial state, observer camera and crop.
  Ordinary columns match action indices; the final column shows each
  trajectory's terminal state.}
  \label{fig:app-vector-sequence}
\end{figure*}

\subsection{Real-robot protocol}
\label{app:real-robot}
\label{app:pi05-integration}

We evaluate Charging Connector Insertion using a $\pi_{0.5}$ policy
on a PiPER platform with fixed external and wrist-mounted RealSense
D455 views. The reported setting uses 10\,Hz action execution and
$K=4$ executed actions per policy query. The seven configurations in
main-paper Table~\ref{tab:real-robot} each contain 50 trials, totaling 350 physical executions.
Continuous execution supplies the unquantized source reference.
Memoryless and CarryQ are compared at 4 and 6 bits; absolute-target
memoryless encoding at both rates supplies a separate representation
control. Bit budgets refer to each encoded continuous coordinate.

The source policy outputs absolute joint targets. At each query,
let $j_0$ be the observed joint state and $j_1,\ldots,j_K$ the targets
selected for execution. The shared incremental representation is
\begin{equation}
 d_1=j_1-j_0,\qquad d_t=j_t-j_{t-1}\quad(t>1).
 \label{eq:piper-increments}
\end{equation}
Both memoryless encoding and CarryQ decode an increment $\hat d_t$;
the same integrating decoder then reconstructs the commanded target,
\begin{equation}
 \hat j_t=j_0+\sum_{s=1}^{t}\hat d_s.
 \label{eq:piper-reconstruction}
\end{equation}
Without quantization, this transformation recovers the source targets.
CarryQ retains residuals across the four executed increments and resets
them at the next policy query. Gripper commands bypass encoding and
carry no residual. Absolute-target memoryless controls instead quantize
$j_t$ directly. Thus the primary comparison changes temporal encoding
within a common incremental interface; the absolute-target rows test
a different representation. These quantities describe joint-target
commands rather than measured end-effector displacements.
The encoding stage follows the $\pi_{0.5}$ flow-based action expert and
output conversion, preserving the backbone, training objective and sampler.
Residuals connect action steps rather than flow-sampling iterations.
Table~\ref{tab:model-integration} contrasts the policy interfaces.

Alignment (S1) denotes reaching the position and orientation required
for insertion. S2 denotes complete insertion with electrical connection,
the same completion criterion used in the simulated insertion tasks.
A measurement device verifies this connection on the real robot.
S2 is a subset of S1 within each execution. Both stage rates divide
success counts by all 50 trials in the corresponding configuration;
S2 is not conditioned on alignment success.
Main Figure~\ref{fig:real-charger-insertion}(c) shows the insertion sequence.

At 4 bits, CarryQ adds 17 alignment successes and 15 final successes
over memoryless encoding; at 6 bits, it adds 15 and 13.
These count differences yield S1 gains of 34 and 30 percentage points,
and S2 gains of 30 and 26 points, respectively.
The absolute-target controls are kept distinct from these same-budget
CarryQ--memoryless differences. The two stages describe outcomes within
the same trial and do not double the physical sample count.

Before downstream safety processing, writing $e_t=\hat d_t-d_t$ yields
\begin{equation}
 \hat j_k-j_k=\sum_{t=1}^{k}e_t,\qquad 1\leq k\leq K.
 \label{eq:joint-reconstruction-error}
\end{equation}
Thus cumulative increment error equals reconstructed joint-target error.
The unchanged safety limiter and joint controller determine subsequent
tracking; no kinematics replacement is introduced. The observed
$j_0\in\mathbb R^6$ is a common reconstruction anchor. Both encoders use
identical increment ranges and six encoded coordinates, costing $6b$
motion bits per step; any communication of the anchor is shared.
The source prediction horizon is preserved, and only its first four
selected targets are encoded.

\section{Replay and Transfer}
\label{app:reproducibility}

\subsection{Replay tasks}
\label{app:replay-tasks}

The six LIBERO-90 tasks cover diverse scenes:
\begin{enumerate}
  \item \texttt{KITCHEN\_SCENE1}: put the black bowl on top of the cabinet;
  \item \texttt{KITCHEN\_SCENE10}: put the butter at the front in the top
  drawer of the cabinet and close it;
  \item \texttt{KITCHEN\_SCENE3}: put the frying pan on the stove;
  \item \texttt{LIVING\_ROOM\_SCENE1}: pick up the alphabet soup and put it in
  the basket;
  \item \texttt{LIVING\_ROOM\_SCENE3}: pick up the ketchup and put it in the
  tray; and
  \item \texttt{STUDY\_SCENE4}: pick up the book in the middle and place it on
  the cabinet shelf.
\end{enumerate}
Screening evaluates all 90 tasks on demonstrations 0--9 using continuous
replay and memoryless encoding at 3, 4 and 5 bits.
Eligibility requires at least eight successful continuous replays,
monotone memoryless success over these rates, and 4-bit success between
20\% and 80\%. Of the 57 eligible tasks, six were selected for scene and
behavior diversity before evaluating CarryQ.
The vector codebooks are fitted on demonstrations 0--9; demonstrations
10--49 supply held-out actions and initial conditions.
Scalar grids are prescribed, and Praxis retains its native bin definitions.
All comparisons share recorded source actions, simulator initial and
model states, and the decoder for the selected interface.

The primary cohort retains 230 of the 240 held-out conditions where
continuous-reference replay succeeds. Scalar and Praxis comparisons
use five rates from 3 to 7 bits; vector comparisons use four rates
from 4 to 7 bits. The primary vector codebook uses fitting seed 0,
with seeds 1 and 2 assessing stability. These are representation
replays driven by recorded commands, rather than learned-policy outputs.

\subsection{Cluster bootstrap}
\label{app:bootstrap}

The estimator averages rates within demonstrations, demonstrations within
tasks, and tasks equally.
Let $y^{\mathrm{CQ}}_{i,j,b}$ and $y^{\mathrm{ML}}_{i,j,b}$ denote binary
success for task $i$, demonstration $j$ and rate $b$.
With $n_i$ retained demonstrations and rate set $\mathcal{B}$, the
percentage-point difference is
\begin{equation}
 \widehat{\delta}=\frac{100}{6}\sum_{i=1}^{6}\frac{1}{n_i}
 \sum_{j=1}^{n_i}\frac{1}{|\mathcal{B}|}
 \sum_{b\in\mathcal{B}}
 \bigl(y^{\mathrm{CQ}}_{i,j,b}-y^{\mathrm{ML}}_{i,j,b}\bigr).
 \label{eq:paired-estimand}
\end{equation}
Reference filtering retains 38, 40, 40, 38, 37 and 37 demonstrations
in task order, with matched task weights. The bootstrap draws 10,000 whole-demonstration samples within task,
retaining paired encoder outcomes and all rates together.
It preserves trajectory dependence and conditions on the fixed cohort.
The implementation records iteration order and uses NumPy \texttt{default\_rng(0)}.
The 95\% gain intervals are [27.2677, 32.6830] pp for scalar,
[5.0715, 12.5865] pp for the primary vector seed, and
[28.4643, 32.8429] pp for native Praxis.

Jointly resampling the five scalar mechanism quantities gives 95\%
reduction intervals of [87.14, 87.49]\% for chunk-summed norm,
[81.23, 81.49]\% for maximum prefix norm and [94.64, 95.06]\%
for rollout-summed norm. Pointwise MSE increases by 40.00\%
[39.28, 40.73]\%, while first-difference MSE increases by
5.45$\times$ [5.35$\times$, 5.55$\times$].

\subsection{Independent task cohort}
\label{app:task-transfer}

The task extension uses a prespecified hash-based selection independent
of CarryQ outcomes, with six kitchen, four living-room and two study tasks.
Each contributes demonstrations 10--19 at 4 and 6 bits.
All 120 conditions are retained, with equal task and rate weights.
The tasks are closing the microwave, stowing pudding, placing the
yellow-and-white mug, stowing the wine bottle, placing the pan and bowl,
placing the red mug, stacking bowls, placing the white mug, placing butter,
and two book-placement tasks.

\begin{table}[tbp]
 \centering
 \caption{Success counts supporting the main execution summary.
Counts and rates use the same executions;
repeated rates and sources do not create independent initial states.
Differences are CarryQ minus memoryless in percentage points.}
\label{tab:aggregate-results}
 \tablestyle
\begin{tabularx}{\columnwidth}{YrQr}
\toprule
\tablehead
Cohort & \makecell[r]{Memoryless\\{\tablesubfont\color{tableaccent} \% (successes)}} & \makecell[r]{\cqhead\\{\tablesubfont\color{tableaccent} \% (successes)}} & $\Delta$ pp \\
\midrule
12 tasks, 4 and 6 bits & \tablepair{58.75}{141/240} & \tablepair{92.92}{223/240} & +34.17 \\
\addlinespace[3pt]
SeedPolicy, six settings & \tablepair{20.83}{25/120} & \tablepair{33.33}{40/120} & +12.50 \\
\addlinespace[3pt]
Can, confirmation & \tablepair{52.25}{209/400} & \tablepair{57.75}{231/400} & +5.50 \\
\addlinespace[3pt]
Can, independent reset & \tablepair{51.75}{207/400} & \tablepair{55.75}{223/400} & +4.00 \\
\bottomrule
\end{tabularx}
\end{table}

\subsection{Clipping incidence}
\label{app:clipping}

A trigger occurs when an adjusted input exceeds the allowed action
range or a scalar residual exceeds its half-grid bound before clipping.
Counts use strict inequalities without numerical tolerance, on encoding
traces whose decoded actions match the replay inputs.
For scalar $H=8$, CarryQ triggers on 2,784 of 1,126,500 eligible
step--coordinate updates (0.247\%); 759 of 1,150 trajectories contain
at least one trigger. The memoryless range path has zero triggers.
For vector CarryQ, input-range clipping occurs on 58,461 of 2,703,600
updates (2.162\%), across 2,642 of 2,760 trajectories.
The memoryless vector path has no corresponding range-clipping operation.
The Praxis-compatible endpoint-grid idealization has 676 triggers in
1,314,250 updates (0.0514\%); these counts concern that idealization
and do not measure clipping in native Praxis.

\subsection{Replay examples}
\label{app:replay-sequences}

Figures~\ref{fig:app-e3-sequence}--\ref{fig:app-vector-sequence}
show how encoding changes execution on a shared recorded source.
The scalar pan example retains the median-completion case among 24
pairs where memoryless encoding fails and CarryQ succeeds: the median
completion step is 180, with nearest-case and smaller-index tie-breaking.
The two native Praxis examples retain the anchors selected before
viewing their renderings. Additional examples were drawn from two
completion-time quantiles within each task-interface pair among positive
paired outcomes before rendering. The vector book example uses fitting
seed 0. These cases illustrate the full paired evaluation.

Each example uses a fixed observer camera and square crop shared across
methods and time. Ordinary columns match action indices; the final
column shows each trajectory's actual terminal state. The continuous
reference row in Figure~\ref{fig:app-vector-sequence} executes the
unquantized source commands. Observer images are rendered from recorded
simulator states with the action sequences unchanged.

\subsection{Code and data}

The analysis package provides paired outcomes, encoder implementations,
configurations and statistical code for the six-task replay and clipping
analyses. Its README records input versions, checksums and reproduction
commands. We use official LIBERO demonstrations and simulator assets.

\section{Temporal Analysis}
\label{app:fine-increment}

Command diagnostics examine amplitude, directional persistence and error
alignment in recorded trajectories. The association analysis uses the
replay outcomes paired with these diagnostic traces; aggregate execution
success is reported separately in Table~\ref{tab:aggregate-results}.

\subsection{Measurement windows}
\label{app:metrics}

Replay command diagnostics include every recorded source action and the
partial final chunk. Chunk-summed norm is the mean norm of the sum within
each chunk; maximum prefix norm records the largest prefix discrepancy;
rollout-summed norm uses the complete recorded sequence.
Decoded-command first-difference MSE measures command variation.
For decomposition, $L$, $X$ and $T$ are computed for each valid chunk
before averaging; mean squared endpoint error is distinct from the
square of a mean endpoint norm. Task-extension prefix-norm AUC divides
the sum of valid prefix norms by valid action count.

These measurements resolve different consequences of the same error
sequence. Local energy accounts for each error independently, while the
endpoint includes all pairwise interactions within the window.
Maximum prefix norm additionally records the largest intermediate
excursion. For example, compensation concentrated at the final action
can improve an endpoint without improving earlier prefixes.
Reporting both quantities tests whether cancellation also develops during
execution. Complete-window analysis covers all command directions and amplitudes
at fixed length.

Resetting leaves issued commands unchanged but clears the residual
available for later compensation. In the unclipped case, the endpoint
residual of each reset window contributes to the full-sequence sum.
Window-level bounds and rollout measurements therefore address different
time scales. Similarly, residual feedback may cross codeword boundaries
to compensate rounding, increasing decoded first differences while
reducing prefix or endpoint error. Temporal accuracy and smoothness
measure different properties of the resulting command sequence.

\subsection{Fine-increment strata}

The cohort contains 480 quantized sequences and 120 continuous sources,
totaling 600 replays over 12 tasks.
Run and persistence analyses use the first three translation-command
coordinates. For source magnitude $\alpha_{t,d}=|a_{t,d}|/\Delta_{b,d}$,
the midpoint grid on $[-1,1]$ has spacing $\Delta_b=2/2^b$.
The endpoint-inclusive spacing is instead $2/(2^b-1)$.
Irregular vector codebooks require local codeword distances rather than
this scalar spacing.

Source-direction runs comprise nonzero, same-sign commands truncated
at $H=8$ boundaries; zeros break runs.
Long runs have 4--8 steps, and mean-small runs have mean $\alpha\leq0.5$.
Error-sign runs are defined separately from the decoded-minus-source
errors; an exact zero error breaks an error-sign run.
The ``one sign'' statistic in main Table~\ref{tab:observed-bias} requires a common strict error sign throughout the run.

Run diagnostics retain the encoder's residual history before the run;
only the measurement window begins at its boundary.
We average runs within demonstrations, demonstrations within tasks and
tasks equally. Table~\ref{tab:fine-increment-strata} gives counts for nonempty strata.
Local energy $L$ sums squared errors, $T$ is squared run-endpoint error,
and prefix magnitude averages absolute cumulative error over the run.

\begin{table}[tbp]
 \centering
 \caption{Translation-command error by amplitude and run length.
 Changes compare CarryQ with memoryless, in percent; arrows indicate
 increases or decreases. Counts give contributing tasks, demonstrations and coordinate runs.
 $L$ is local squared-error energy; $T$ is squared run-endpoint error.
 Prefix is the mean absolute cumulative error within the run.}
 \label{tab:fine-increment-strata}
 \tablestyle
\begin{tabularx}{\columnwidth}{Yrrr}
\toprule
\tablehead
\makecell[l]{Amplitude, length\\{\tablesubfont\color{tableaccent} Tasks; demos; runs}} & $\Delta L$ (\%) & $\Delta T$ (\%) & \makecell[r]{$\Delta$ prefix\\(\%)} \\
\midrule
\tablegroup{4}{4-bit encoding}
\tablelabel{Small, long}{12; 116; 617} & $\uparrow$109.11 & $\downarrow$91.55 & $\downarrow$56.04 \\
\tablelabel{Small, short}{12; 120; 1044} & $\uparrow$55.50 & $\downarrow$44.17 & $\downarrow$8.75 \\
\tablelabel{Larger, long}{12; 120; 5685} & $\uparrow$90.32 & $\downarrow$88.12 & $\downarrow$47.58 \\
\tablelabel{Larger, short}{12; 118; 612} & $\uparrow$103.01 & $\downarrow$15.73 & $\uparrow$4.95 \\
\addlinespace[2pt]
\tablegroup{4}{6-bit encoding}
\tablelabel{Small, long}{11; 38; 46} & $\uparrow$123.84 & $\downarrow$81.72 & $\downarrow$37.62 \\
\tablelabel{Small, short}{12; 111; 344} & $\uparrow$91.38 & $\downarrow$4.48 & $\uparrow$13.89 \\
\tablelabel{Larger, long}{12; 120; 6256} & $\uparrow$85.56 & $\downarrow$87.63 & $\downarrow$46.30 \\
\tablelabel{Larger, short}{12; 120; 1312} & $\uparrow$88.07 & $\downarrow$20.19 & $\uparrow$4.63 \\
\bottomrule
\end{tabularx}
\end{table}

The source-only exposure descriptor measures the fraction of recorded
translation coordinates in small, long runs. Its mean is 6.302\% at
4 bits and 0.413\% at 6 bits.
Benefits also occur in larger long runs, while some short-run strata
have increased prefix error. This variation distinguishes amplitude from the duration of error accumulation.

Strict sub-bin runs impose the stronger condition that every command
has magnitude strictly between zero and $\Delta_b/2$.
The midpoint decoder outputs $\operatorname{sign}(a_t)\Delta_b/2$,
so memoryless error follows the source sign.
At 4 bits, 528 long runs span 114 demonstrations and 12 tasks;
mean endpoint magnitude falls from 1.342 to 0.322 grid units.
At 6 bits, 25 runs span 19 demonstrations and nine tasks;
the corresponding change is from 1.072 to 0.405 grid units.
These runs occupy 4.878\% and 0.194\% of translation step--coordinates
at the respective budgets, using equal demonstration and task weights.

\begin{figure*}[t]
  \centering
  \begin{minipage}[t]{0.477\textwidth}
    \noindent{\small a}\par\vspace{1pt}
    \includegraphics[width=\linewidth]{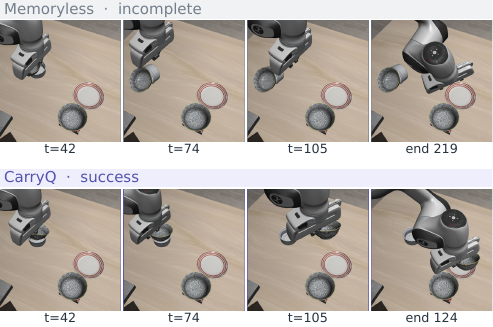}
  \end{minipage}\hfill
  \begin{minipage}[t]{0.477\textwidth}
    \noindent{\small b}\par\vspace{1pt}
    \includegraphics[width=\linewidth]{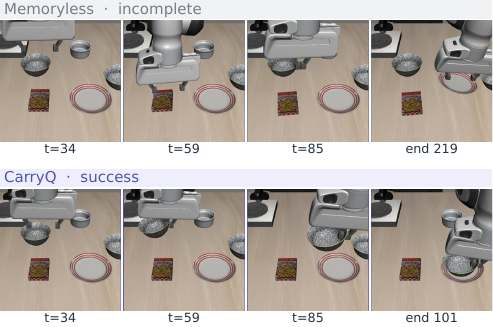}
  \end{minipage}
  \caption{Closed-loop OpenVLA-OFT executions under camera (a) and sensor
  (b) perturbations at 4 bits. Memoryless encoding leaves the target bowl
  tilted away from the plate (a) or beside it (b), while CarryQ reaches the
  goal. Common observer views show the saved executions; panel (b) uses
  uncorrupted observer images of actions generated under the original
  sensor perturbation. Ordinary columns match action indices, and each
  final column shows the two terminal states.}
  \label{fig:app-online-sequences}
\end{figure*}

\begin{figure*}[t]
  \centering
  \begin{minipage}[t]{0.477\textwidth}
    {\small a}\par\vspace{1pt}
    \includegraphics[width=\linewidth]{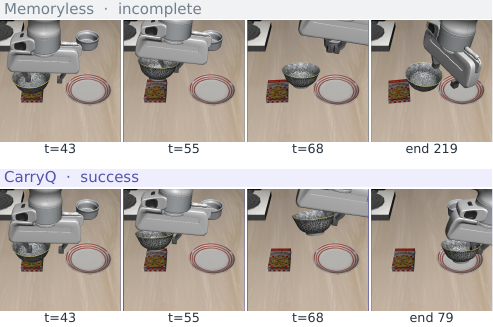}
  \end{minipage}\hfill
  \begin{minipage}[t]{0.477\textwidth}
    {\small b}\par\vspace{1pt}
    \includegraphics[width=\linewidth]{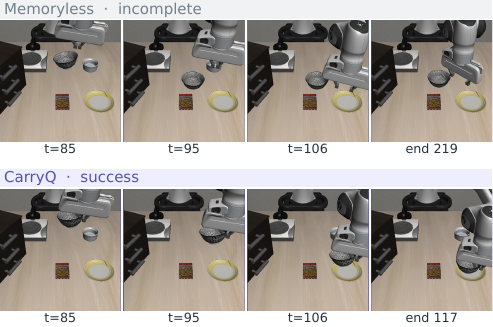}
  \end{minipage}
  \caption{OpenVLA-OFT sequences on Spatial (a) and Pro (b) at 4 bits.
  CarryQ places the bowl at the goal, while memoryless encoding leaves it
  at the source. Each pair uses its own closed-loop commands and a shared
  initial state, camera and crop. Ordinary columns match action indices;
  the final column shows each arm's terminal state.}
  \label{fig:app-spatial-pro-sequences}
\end{figure*}

\begin{figure*}[t]
  \centering
  \begin{minipage}[t]{0.477\textwidth}
    {\small a}\par\vspace{1pt}
    \includegraphics[width=\linewidth]{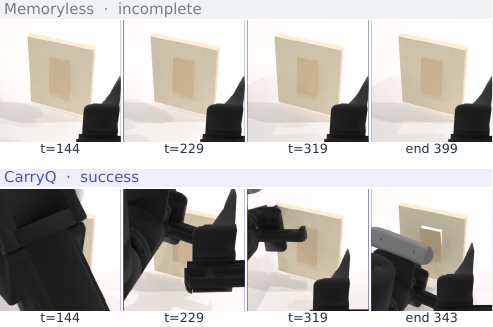}
  \end{minipage}\hfill
  \begin{minipage}[t]{0.477\textwidth}
    {\small b}\par\vspace{1pt}
    \includegraphics[width=\linewidth]{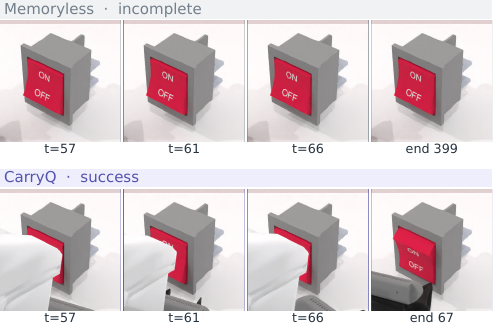}
  \end{minipage}
  \caption{RoboTwin switch turning with SeedPolicy at 4 bits.
  Two paired executions with different native switch assets show CarryQ
  completing the turn while memoryless encoding remains unsuccessful.
  Camera and crop are shared. Ordinary columns match action indices;
  the last shows the two terminal states.}
  \label{fig:app-robotwin-sequences}
\end{figure*}

\begin{figure*}[t]
  \centering
  \begin{minipage}[t]{0.477\textwidth}
    {\small a}\par\vspace{1pt}
    \includegraphics[width=\linewidth]{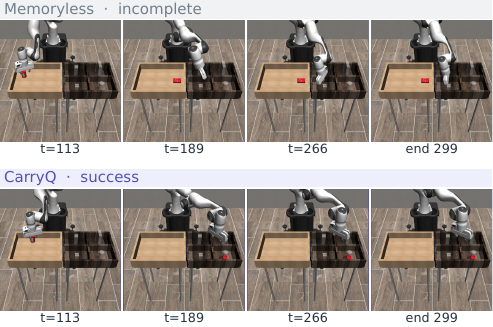}
  \end{minipage}\hfill
  \begin{minipage}[t]{0.477\textwidth}
    {\small b}\par\vspace{1pt}
    \includegraphics[width=\linewidth]{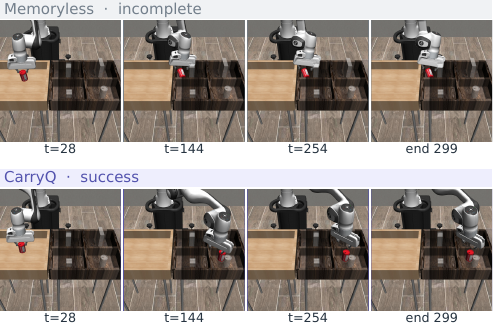}
  \end{minipage}
  \caption{Robomimic Can under pretrained (a) and DSRL-adapted (b)
  policies at 4 bits. CarryQ completes placement in both selected pairs.
  Shared observer views render the saved simulator states without
  changing actions. Ordinary columns match action indices; the final
  column shows each arm's terminal state.}
  \label{fig:app-can-sequences}
\end{figure*}

\subsection{Bias and task outcomes}
\label{app:bias-outcomes}
\label{app:complete-windows}

Complete-window diagnostics cover the full normalized XYZ source
sequences, including tails beyond an episode's early success.
The sources contain 19,418 steps and 9,427 exact-zero translation
coordinates, all within $[-1,1]$.
Zeros remain in this analysis, unlike the source-direction run extraction.
Each budget provides 7,125 nonoverlapping eight-step coordinate windows.
Only the 1,254 trailing step--coordinates outside complete windows are
omitted from these summaries; every demonstration outcome is retained.

Endpoint magnitude is $B=|\sum_t e_t|/\Delta_b$; mean prefix magnitude
is $\operatorname{mean}_k|\sum_{t\leq k}e_t|/\Delta_b$.
Error coherence is $|\sum_t e_t|/\sum_t|e_t|$, assigned zero when
its denominator is zero. Complete-window persistence means an error-sign
run of at least four steps within a window.
Window summaries average within demonstrations, within tasks and then
equally across tasks. Main Table~\ref{tab:bias-window-ci} reports the resulting encoder means.
Its exploratory intervals use 5,000 whole-demonstration resamples within
task with seed 20261113, retaining both rates and encoders together.

The persistence distribution uses each demonstration's mean across
complete windows. The joint plot divides CarryQ mean local energy and
mean squared endpoint error by their memoryless counterparts within
each demonstration and rate.
All 240 pairs have larger local energy and smaller squared endpoint
error; both rates use the same sources.

Success associations pair each diagnostic replay's binary encoder
outcomes, center both the descriptor and benefit within task, and
compute Pearson correlation. Whole demonstrations are resampled within
task, carrying rates and encoder variants together.
Small-long exposure has correlations 0.013 and 0.028 at 4 and 6 bits.
Table~\ref{tab:bias-success-association} reports all six correlations
and unadjusted intervals.

\begin{table}[tbp]
 \centering
 \caption{Exploratory associations with paired binary task benefit
 (CarryQ minus memoryless), centered within task, using 120 demonstrations
 per budget. Intervals resample whole demonstrations within task and
 are unadjusted for multiple comparisons.}
 \label{tab:bias-success-association}
 \tablestyle
\begin{tabularx}{\columnwidth}{Yrr}
\toprule
\tablehead
Episode descriptor & \shortstack[r]{\strut 4-bit $r$\\{\tablesubfont\color{tableaccent} [95\% CI]}} & \shortstack[r]{\strut 6-bit $r$\\{\tablesubfont\color{tableaccent} [95\% CI]}} \\
\midrule
Small nonzero step fraction & \tablepair{0.031}{[$-$0.146, 0.203]} & \tablepair{0.086}{[$-$0.075, 0.238]} \\
\addlinespace[3pt]
Mean-small long-run exposure & \tablepair{0.013}{[$-$0.154, 0.166]} & \tablepair{0.028}{[$-$0.129, 0.173]} \\
\addlinespace[3pt]
Strict sub-bin long-run exposure & \tablepair{0.012}{[$-$0.151, 0.166]} & \tablepair{0.034}{[$-$0.065, 0.150]} \\
\addlinespace[3pt]
ML persistent-error-window fraction & \tablepair{0.040}{[$-$0.178, 0.252]} & \tablepair{0.091}{[$-$0.080, 0.256]} \\
\addlinespace[3pt]
ML mean endpoint magnitude & \tablepair{0.055}{[$-$0.111, 0.216]} & \tablepair{0.206}{[0.029, 0.356]} \\
\addlinespace[3pt]
ML mean prefix magnitude & \tablepair{0.053}{[$-$0.105, 0.210]} & \tablepair{0.193}{[0.023, 0.352]} \\
\bottomrule
\end{tabularx}
\end{table}

Outcome-group means describe the diagnostic replays underlying these
associations. Table~\ref{tab:outcome-means} aligns counts $n$ and mean memoryless
endpoint magnitude $B$ at each precision:

\begin{table}[tbp]
\centering
\caption{Outcome-group counts and mean memoryless endpoint magnitude in grid units.}
\label{tab:outcome-means}
\tablestyle
\begin{tabularx}{\columnwidth}{Yrrrr}
\toprule
\tablehead
Outcome & $n_4$ & $B_4$ & $n_6$ & $B_6$ \\
\midrule
Rescued & 67 & 1.177 & 16 & 1.208 \\
Regressed & 2 & 1.044 & 5 & 1.099 \\
Both succeed & 41 & 1.131 & 93 & 1.141 \\
Both fail & 10 & 1.152 & 6 & 1.170 \\
\bottomrule
\end{tabularx}
\end{table}

Small-run exposure is weakly associated with task benefit at both
budgets; cumulative bias has a positive exploratory association at 6 bits.
Source descriptors include the complete recorded trajectories.

\subsection{Error decomposition}
\label{app:decomposition}

The six-coordinate decomposition uses valid $H=8$ chunks, including
partial final chunks, from the 12-task command traces.
Each chunk's local energy $L$, cross-time term $X$ and squared endpoint
$T$ are computed before averaging. Table~\ref{tab:error-components} gives the rounded means.

\begin{table}[tbp]
\centering
\caption{Local energy, cross-time interaction and squared endpoint error on the 12-task traces.}
\label{tab:error-components}
\tablestyle
\begin{tabularx}{\columnwidth}{Yrrr}
\toprule
\tablehead
Encoding & $L$ & $X$ & $T$ \\
\midrule
\tablegroup{4}{4-bit encoding}
Memoryless & 0.09658 & +0.33692 & 0.43350 \\
CarryQ & 0.13777 & $-$0.13110 & 0.00667 \\
\tablegroup{4}{6-bit encoding}
Memoryless & 0.005943 & +0.016570 & 0.022514 \\
CarryQ & 0.008486 & $-$0.008046 & 0.000439 \\
\bottomrule
\end{tabularx}
\end{table}

The squared endpoint reductions are 98.46\% at 4 bits and 98.05\%
at 6 bits. Prefix-norm AUC decreases by 76.85\% and 74.49\%,
respectively, using the valid-action average defined in the main paper.
The chunk-level identity $L+X=T$ holds to a maximum numerical residual
of $4.44\times10^{-16}$ before rounding.

This distinction also matters when two sequences both cancel at their
endpoints. Placing all positive errors before all negative errors creates
a large intermediate prefix, whereas alternating their signs keeps each
prefix small. Both orderings can have identical local energy and zero
endpoint error. The maximum prefix norm captures the largest accumulated
command discrepancy encountered along the window; the endpoint norm
records only what remains after all actions have contributed.
These complementary summaries distinguish temporal cancellation from
its timing within an execution window.

For a matrix view, stack $e_t^\top$ into $E\in\mathbb R^{H\times D}$
and let $C$ be the lower-triangular prefix-sum operator. Then
\begin{equation}
 P(E)=CES^\top,
 \label{eq:prefix-matrix}
\end{equation}
whose rows contain the cumulative discrepancies at successive prefixes.
For constant errors $e_t=\epsilon$ and alternating errors
$e_t=(-1)^t\epsilon$ over an even-length window, both local energies are
$H\epsilon^2$, but squared endpoints are $H^2\epsilon^2$ and zero.
Their cross-time terms are $H(H-1)\epsilon^2$ and $-H\epsilon^2$.
Nearest-codeword selection minimizes independent local costs and omits
these cross-products; a residual makes the earlier discrepancy available
to subsequent token choices.

\section{Closed-Loop Evaluation}
\label{app:closed-loop}
\label{app:closeout}

Table~\ref{tab:deployment} gives native controls for settings with an
unquantized arm. Each comparison retains its policy checkpoint and
controller, starts from paired initial states and subsequently replans
from each arm's own observations.

\begin{table}[tbp]
 \centering
 \caption{Continuous-policy controls for cohorts with an unquantized arm: success (\%).
Native denotes continuous source-policy execution; gains compare CarryQ
with memoryless (ML). The independent reset cohort has no continuous arm
and is reported separately in Table~\ref{tab:can-reset}.}
 \label{tab:deployment}
 \tablestyle
\begin{tabularx}{\columnwidth}{YrrQr}
\toprule
\tablehead
Environment & Native & ML & \cqhead & $\Delta$ pp \\
\midrule
\tablegroup{5}{OpenVLA-OFT}
Spatial & 100.00 & 98.75 & 100.00 & +1.25 \\
Pro Spatial & 98.75 & 95.00 & 97.50 & +2.50 \\
Plus Spatial & 80.00 & 78.75 & 83.75 & +5.00 \\
\addlinespace[2pt]
\tablegroup{5}{SeedPolicy}
Six settings & 27.50 & 20.83 & 33.33 & +12.50 \\
\addlinespace[2pt]
\tablegroup{5}{Can}
Confirmation & 53.50 & 52.25 & 57.75 & +5.50 \\
\bottomrule
\end{tabularx}
\end{table}

\subsection{OpenVLA-OFT}
\label{app:oft-evaluation}

Online evaluations pair initial states and subsequently let each arm
replan on its own observations. OpenVLA-OFT uses 80 states in each of
three environments, totaling 720 rollouts across the continuous,
memoryless and CarryQ arms. Paired 95\% gain intervals are [0, 3.75],
[$-$2.5, 7.5] and [$-$3.75, 13.75] pp for Spatial, Pro Spatial and
Plus Spatial. Pairing uses initial states; all are retained, including two
Plus states whose initial predictions differ across arms.

Figures~\ref{fig:app-online-sequences} and~\ref{fig:app-spatial-pro-sequences}
extend the main paper's LIBERO outcomes with camera, sensor and task variations.
The selected positive pairs use the saved actions from the original
closed-loop evaluations, with each arm retaining its own policy queries
and commands. Replaying these actions from matched initial states
reproduces their recorded success status and ending action counts,
providing a common high-resolution observer view. Sensor examples use
a clean observer of the original perturbed executions.

\subsection{RoboTwin}
\label{app:seedpolicy}

SeedPolicy on RoboTwin contributes six settings with 20 initial states each.
Its interface uses absolute joint targets followed by TOPP and
drive-target control. Table~\ref{tab:seedpolicy-complete} gives setting-level controls
and the CarryQ aggregate.

\begin{table}[tbp]
 \centering
 \caption{SeedPolicy controls: successes out of 20. Native and memoryless
results are shown by setting. CarryQ's aggregate total is 40/120 (33.33\%).
Actions are absolute joint targets.}
\label{tab:seedpolicy-complete}
 \tablestyle
\begin{tabularx}{\columnwidth}{Ylrr}
\toprule
\tablehead
Task & Setting & Native & Memoryless \\
\midrule
Cup & Clean & 3 & 1 \\
& Randomized & 0 & 0 \\
\addlinespace[3pt]
Bowls & Clean & 13 & 6 \\
& Randomized & 0 & 0 \\
\addlinespace[3pt]
Switch & Clean & 14 & 9 \\
& Randomized & 3 & 9 \\
\midrule
\multicolumn{2}{l}{Total (out of 120)} & 33 & 25 \\
\bottomrule
\end{tabularx}
\end{table}

The SeedPolicy command diagnostic covers full six-action predictions
with 12 eligible coordinates per action. Local energy is therefore
$L=72\,\mathrm{MSE}$, and the per-query summaries recover
$X=T-L$ from squared chunk-sum error $T$.
The 120 first-query sources are shared across arms; subsequent queries
may diverge. Of the 360 diagnostic episodes, 247 have an unexecuted
tail in the final prediction. These diagnostics measure predicted joint commands.

Figures~\ref{fig:app-can-sequences} and~\ref{fig:app-robotwin-sequences}
show the corresponding transport and articulated-control outcomes
beyond LIBERO. Can images render saved paired executions; RoboTwin
images come from additional paired recordings with the same checkpoint,
encoder and controller. These selected qualitative examples are separate
from the aggregate cohorts in Table~\ref{tab:aggregate-results}.

\subsection{Can}
\label{app:can-confirmation}

The robomimic Can task uses low-dimensional observations with Native and DSRL-adapted
source checkpoints. Each cohort has 100 initial states shared across
two sources and two bit budgets, giving 400 executions per encoding arm.
An earlier 440-rollout development set is excluded from both cohorts.
DSRL's separate three-training-seed comparison raises continuous success
from 33\% to 49\% on 200 shared states. That comparison measures source
adaptation, whereas CarryQ comparisons hold each source fixed.

\begin{table}[tbp]
 \centering
 \caption{Can confirmation controls: success (\%), 100 shared states per
source and rate cell. RR averages three random-rounding seeds. CarryQ's
aggregate is 231/400 (57.75\%).}
\label{tab:can-confirmation-controls}
 \tablestyle
\begin{tabularx}{\columnwidth}{Yrrr}
\toprule
\tablehead
Source and rate & Continuous & Memoryless & RR mean \\
\midrule
\tablegroup{4}{Native source}
4-bit & 43 & 45 & 46.00 \\
6-bit & 43 & 46 & 43.00 \\
\addlinespace[2pt]
\tablegroup{4}{DSRL source}
4-bit & 64 & 55 & 56.00 \\
6-bit & 64 & 63 & 59.33 \\
\bottomrule
\end{tabularx}
\end{table}

The confirmation study includes 2,200 rollouts across continuous,
memoryless, CarryQ and random-rounding arms. Random-rounding means
(Table~\ref{tab:can-confirmation-controls}) average three seeds on 100 matched states
(Table~\ref{tab:can-rounding-seeds}).

\begin{table}[tbp]
\centering
\caption{Can random-rounding success by seed (\%).}
\label{tab:can-rounding-seeds}
\tablestyle
\begin{tabularx}{\columnwidth}{Yrrr}
\toprule
\tablehead
Source and rate & \multicolumn{3}{r}{Seed runs (\%)} \\
\midrule
Native, 4-bit & 47 & 45 & 46 \\
Native, 6-bit & 42 & 44 & 43 \\
DSRL, 4-bit & 54 & 61 & 53 \\
DSRL, 6-bit & 59 & 61 & 58 \\
\bottomrule
\end{tabularx}
\end{table}

Can diagnostics use unnormalized controller commands, including translation
scale and input saturation. Prefix AUC sums four prefix norms per query
in meter-steps. The task-extension measure instead divides by valid action
count; dividing Can AUC by fixed $K=4$ preserves every relative change.
Main Table~\ref{tab:robustness-consistency} gives reductions by source and rate across 400 paired states (task outcomes: Table~\ref{tab:aggregate-results}).

\section{Reset and Learning}
\label{app:reset-learning}

\subsection{Residual lifetime}
\label{app:reset-cohort}

Reset comparisons keep checkpoints, bit budgets and the four-action
execution window fixed. The two-step variant clears feedback halfway
through each query; CarryQ retains it for all four actions.
Main Figure~\ref{fig:ef-main}(d) re-encodes 200 stored continuous-policy trajectories,
each comprising 75 queries of four actions. Two fixed policy sources,
two bit budgets and 100 matched initial states yield 400 source--rate--state
conditions. Prefix AUC sums the four translation-command error prefix
norms within each query, including across a two-action residual reset.
The curves average all conditions equally and retain all 75 query values
without smoothing. These fixed-source diagnostics complement the
independent closed-loop comparison below.

The independent closed-loop cohort uses 100 new initial states with both sources
at 4 and 6 bits. Its 1,200 rollouts exclude 12 pilot episodes
and cover memoryless encoding and both reset variants.
Table~\ref{tab:can-reset} gives the memoryless and two-step controls.
The two-step cells sum to 185/400 successes (46.25\%). Four-step
CarryQ gains 4.00 points: 223/400 (55.75\%) versus memoryless
207/400 (51.75\%).
At Native 4-bit, two-step reset loses 11 points against memoryless
(95\% initial-state bootstrap interval [$-$18, $-$5];
20,000 resamples, seed 20261109).

\begin{table}[tbp]
 \centering
 \caption{Independent Can reset controls: success (\%) on 100
shared states per row; four-step CarryQ averages 55.75\%.}
\label{tab:can-reset}
 \tablestyle
\begin{tabularx}{\columnwidth}{Yrr}
\toprule
\tablehead
Source and rate & Memoryless & 2-step reset \\
\midrule
\tablegroup{3}{Native source}
4-bit & 49 & 38 \\
6-bit & 43 & 42 \\
\addlinespace[2pt]
\tablegroup{3}{DSRL source}
4-bit & 62 & 49 \\
6-bit & 53 & 56 \\
\bottomrule
\end{tabularx}
\end{table}

\subsection{Learning encoded targets}
\label{app:encoded-target-learning}

Learning encoded targets poses a different question from encoding a
fixed policy's output.
We also train BAR-Smol, a Praxis blockwise-autoregressive policy with a
SmolVLM backbone, on CarryQ labels.
Across three seeds and 600 arm rollouts, mean success changes from
1.33\% to 7.33\%. The seed-wise effects are +18, 0 and 0 points; the mean gain is +6.00 points
(hierarchical 95\% interval [$-$0.67, 17.67]).

CarryQ constructs training tokens from the continuous demonstration
actions. Matched BAR-Smol policies retain the SmolVLM backbone family,
blockwise prediction schedule, vocabulary and decoder. At inference,
predicted tokens are decoded normally; the policy does not require
reference actions to reconstruct an encoding residual. This evaluation
therefore addresses learning encoded targets separately from encoding
the output of a frozen policy.

\end{document}